\documentclass[11pt]{article}

\usepackage[preprint]{acl}

\usepackage{amsmath}
\usepackage{amssymb}
\usepackage{empheq}

\usepackage[utf8]{inputenc}
\usepackage{float}
\usepackage{latexsym}
\usepackage{cleveref}
\usepackage{array}  
\usepackage{makecell}
\usepackage{booktabs}
\usepackage[most]{tcolorbox}
\usepackage{dirtytalk} 

\usepackage{tikz}
\usetikzlibrary{positioning, patterns, decorations.pathreplacing, calc}

\usepackage{csquotes}
\usepackage{hyperref}
\usepackage{multirow}

\usepackage{fontspec}
\newfontfamily\notoserif[Extension=.ttf, Path=fonts/]{NotoSerif-Regular}

\usepackage[main=english,bidi=default]{babel} 
\babelfont{rm}{TeXGyreTermesX} 

\babelfont[chinese-simplified]{rm}[Extension=.ttf, Path=fonts/]{NotoSerifSC-Regular}
\newfontface\adlamfont[Extension=.ttf, Path=fonts/]{NotoSansAdlam-Regular}
\newfontface\javanesefont[Extension=.ttf, Path=fonts/]{NotoSansJavanese-Regular}
\newfontface\kannadafont[Extension=.ttf, Path=fonts/]{NotoSerifKannada-Regular}
\newfontface\khmerfont[Extension=.ttf, Path=fonts/]{NotoSerifKhmer-Regular}
\newfontface\persoarabicfont[Extension=.ttf, Path=fonts/]{ScheherazadeNew-Regular}
\newfontface\laofont[Extension=.ttf, Path=fonts/]{NotoSerifLao-Regular}
\newfontface\malayalamfont[Extension=.ttf, Path=fonts/]{NotoSerifMalayalam-Regular}
\newfontface\myanmarfont[Extension=.ttf, Path=fonts/]{NotoSerifMyanmar-Regular}
\newfontface\nkofont[Extension=.ttf, Path=fonts/]{NotoSansNKo-Regular}
\newfontface\olfont[Extension=.ttf, Path=fonts/]{NotoSansOlChiki-Regular}
\newfontface\osmanyafont[Extension=.ttf, Path=fonts/]{NotoSansOsmanya-Regular}
\newfontface\odiyafont[Extension=.ttf, Path=fonts/]{NotoSansOriya-Regular}
\newfontface\thaifont[Extension=.ttf, Path=fonts/]{NotoSerifThai-Regular}
\newfontface\telugufont[Extension=.ttf, Path=fonts/]{NotoSerifTelugu-Regular}
\newfontface\tamilfont[Extension=.ttf, Path=fonts/]{NotoSerifTamil-Regular}
\newfontface\bengalifont[Extension=.ttf, Path=fonts/]{NotoSerifBengali-Regular}
\newfontface\hindifont[Extension=.ttf, Path=fonts/]{NotoSerifDevanagari-Regular}
\newfontface\gujaratifont[Extension=.ttf, Path=fonts/]{NotoSerifGujarati-Regular}
\newfontface\balinesefont[Extension=.ttf, Path=fonts/]{NotoSerifBalinese-Regular}

\usepackage{url}
\usepackage{xcolor}
\usepackage{setspace}
\usepackage{bm}
\usepackage{colortbl}
\usepackage{pifont}

\definecolor{darkgray}{RGB}{64, 64, 62} 
\definecolor{rust}{RGB}{204, 120, 92}  
\definecolor{cloudlight}{RGB}{214, 214, 214}
\definecolor{yellow}{RGB}{235, 219, 188}
\definecolor{light_orange}{RGB}{212, 162, 127}
\definecolor{ivory}{RGB}{240,240,235}
\definecolor{red}{RGB}{191, 77, 67}

\newcommand{\highlight}[1]{\colorbox{yellow}{\textbf{#1}}}

\newcommand{\token}[1]{\colorbox{yellow}{\ttfamily\textbf{#1}}\textcolor{white}{\hspace{0.75mm}\rule[-0.18cm]{1.2pt}{0.5cm}\hspace{0.75mm}}}

\usepackage{tikz}

\newcommand{\unknowntokene}[1]{%
  \colorbox{yellow}{\notoserif{\textvisiblespace}#1}%
  \textcolor{white}{\hspace{0.75mm}\rule[-0.18cm]{1.2pt}{0.53cm}\hspace{0.75mm}}%
  }%

\usepackage{microtype}

\usepackage{hyperref}
\usepackage{cleveref}

\usepackage{courier}

\usepackage{graphicx}

\newtcolorbox{CodeBox}[1][]{
    colback=ivory,                
    colframe=black,               
    enhanced jigsaw,
    boxrule=1pt,
    arc=4mm,
    width=\linewidth,
    sharp corners=all,           
    attach boxed title to top center={yshift=-2mm},
    boxed title style={colback=darkgray, colframe=black, fontupper=\bfseries\color{white}},
    title=#1,
}

\newcommand{\Role}[2]{\textbf{\textcolor{darkgray}{\textbf{Role:  }}}\textcolor{rust}{`\hspace{0.5mm}#1\hspace{0.5mm}'}\\\textbf{\textcolor{darkgray}{\textbf{Content: }}}\textcolor{rust}{`\hspace{0.5mm}#2\hspace{0.5mm}'}\\[0.7em]}

\newcommand{\ExpectedOutput}[1]{\textbf{\textcolor{darkgray}{\textbf{Expected Output: }}} #1}

\title{1,729 vs. {\hindifont १७२९}: The Effect of Scripts and Formats on LLM Numeracy}

\author{
 \textbf{Varshini Reddy}\textsuperscript{\dag} \textrm{,}
 \textbf{Craig W. Schmidt}\textsuperscript{\dag} \textrm{,}
 \textbf{Seth Ebner}\textsuperscript{\dag} \textrm{,}
 \textbf{Adam Wiemerslage}\textsuperscript{\dag}\textrm{,}
 \\
 \textbf{Yuval Pinter}\textsuperscript{\S} \textrm{,}
 \textbf{Chris Tanner}\textsuperscript{\dag,\P}
 \\
\begin{tabular}{ccc}
      \textsuperscript{\dag}Kensho Technologies & \textsuperscript{\S}Ben-Gurion University &  \textsuperscript{\P}MIT \\
     Cambridge, MA  & Beer Sheva, Israel &  Cambridge, MA \\
\end{tabular} \\
\texttt{\small varshini.bogolu@kensho.com}
 }

\begin{document}
\maketitle
\begin{abstract}
Large language models (LLMs) have achieved impressive proficiency in basic arithmetic, rivaling human-level performance on standard numerical tasks. However, little attention has been given to how these models perform when numerical expressions deviate from the prevailing conventions present in their training corpora. In this work, we investigate numerical reasoning across a wide range of numeral scripts and formats. We show that LLM accuracy drops substantially when numerical inputs are rendered in underrepresented scripts or formats, despite the underlying mathematical reasoning being identical. We further demonstrate that targeted prompting strategies, such as few-shot prompting and explicit numeral mapping, can greatly narrow this gap. Our findings highlight an overlooked challenge in multilingual numerical reasoning and provide actionable insights for working with LLMs to reliably interpret, manipulate, and generate numbers across diverse numeral scripts and formatting styles.
\end{abstract}

\section{Introduction}
Language models have become foundational to many NLP systems, demonstrating strong performance across tasks that require understanding, generating, and reasoning over text. Because these models are trained and evaluated using natural language inputs and outputs, they are often assessed on their ability to carry out non-linguistic reasoning—such as basic arithmetic—when such problems are expressed in natural language. Numerical reasoning has attracted sustained interest, as it probes whether models can manipulate structured, symbolic information rather than relying solely on surface-level pattern matching. Prior work has shown that LLMs perform arithmetic with high accuracy under controlled settings, with numbers expressed using standard Hindu–Arabic numerals and familiar formatting conventions \citep{wallace-etal-2019-nlp, mccoy-etal-2019-right}.

\begin{figure}
    \centering
    \includegraphics[width=0.975\linewidth]{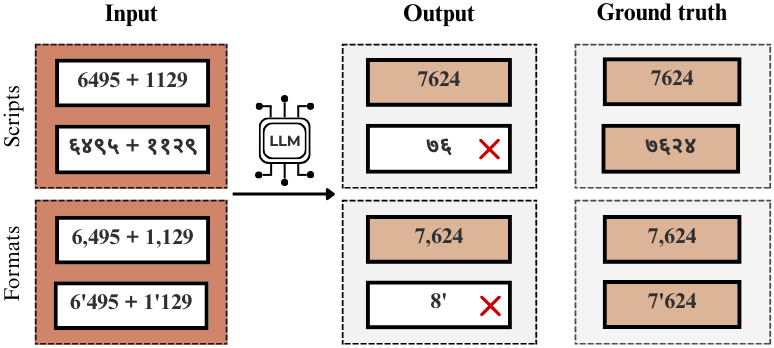}
    \caption{LLM numerical reasoning is sensitive to the script and format used; illustrative samples of inputs and outputs from Llama~3.3.}
    \label{fig:header}
\end{figure}

Despite this success, LLM performance is often contingent on the way information is represented \citep{bhattacharya-etal-2025-investigating, singha2023tabularrepresentationnoisyoperators}. In particular, numbers can appear in diverse scripts and formatting conventions across different languages and regions. Such variations, ranging from alternate numeral scripts to different decimal markers and digit grouping, may introduce challenges for models that have been primarily trained on English-centric corpora. As a result, it remains unclear to what extent previously observed arithmetic competence reflects robust numerical reasoning versus reasoning that is strongly conditioned on well-supported representations.

Tokenization further complicates this picture. Tokenizers are a critical component of LLMs, as they determine how text is segmented into subword units that the model represents and manipulates internally. For numerals in particular, the way a number is tokenized can significantly influence arithmetic performance \citep{singh2024tokenizationcountsimpacttokenization, nogueira2021investigatinglimitationstransformerssimple, zhang2025tokenizationconstraintsllmsstudy}. Numbers represented using fewer, larger tokens are generally easier for models to process, whereas those fragmented into many small tokens introduce additional complexity and increase the likelihood of errors, especially when such token patterns are sparsely observed during training.

In this work, we study numerical reasoning in LLMs by varying only the script and formatting of numerals, while holding the arithmetic itself constant. This allows us to isolate the effect of numeral representation on model performance. Our evaluation probes multiple types of numerical processing, ranging from recognizing and translating numerals to performing arithmetic over non-standard representations.
\Cref{sec:exps} provides an overview of the experimental setup.
In \Cref{sec:numeric_reasoning}, we examine how changes in numeral script affect model behavior, while \Cref{sec:arithmetic_reasoning} analyzes the impact of alternative numeric formatting conventions on arithmetic performance. By focusing on numeral scripts and formatting, an aspect of multilinguality that has received comparatively little attention, this work contributes to a more comprehensive understanding of numerical reasoning in LLMs and highlights the need for evaluation practices that better reflect the diversity of numeric representations encountered in real-world language use.

\section{Approach}
\label{sec:exps}

To study script and format effects on numerical reasoning, we introduce an evaluation dataset consisting of 336 base arithmetic expressions using Hindu–Arabic (HA) digits, with 1–8 digit integer and decimal operands.


\subsection{Multiscript Numeral Dataset}
\label{sec:script_dataset}

Each of the 336 base arithmetic expressions are translated into \textbf{20 distinct numeral scripts}, depicted in \Cref{tab:script_variants}, spanning both high and low-resource scripts. This setup allows us to examine how changes in numeral glyphs affect arithmetic reasoning. Throughout this work, the HA numeral system is treated as the default setting, as it is the standard numeric representation in English and many other languages, and is widely used in text corpora commonly employed to train LLMs.

\begingroup
\setlength{\tabcolsep}{2pt} 
\renewcommand{\arraystretch}{1} 

\begin{table}[h]
\centering
\captionsetup{
    skip=8pt,        
    font={stretch=1.2} 
}

\resizebox{\linewidth}{!}{%
\begin{tabular}{l|cccccccccc|c}
\toprule

\textbf{Script} 
& \multicolumn{10}{c|}{\textbf{Digits}} 
& \textbf{\% Data} \\

\midrule

Hindu-Arabic & 0 & 1 & 2 & 3 & 4 & 5 & 6 & 7 & 8 & 9 & 72.4\\

Perso-Arabic &\persoarabicfont \char"06F0 & \persoarabicfont \char"06F1 & \persoarabicfont \char"06F2 & \persoarabicfont \char"06F3 & \persoarabicfont \char"06F4 & \persoarabicfont \char"06F5 & \persoarabicfont \char"06F6 & \persoarabicfont \char"06F7 & \persoarabicfont \char"06F8 & \persoarabicfont \char"06F9 & 12.2\\

Devanagari & \hindifont ० & \hindifont १ & \hindifont २ & \hindifont ३ & \hindifont ४ & \hindifont ५ & \hindifont ६ & \hindifont ७ & \hindifont ८ & \hindifont ९ & 9.1\\ 

Bengali & \bengalifont ০ & \bengalifont ১ & \bengalifont ২ & \bengalifont ৩ & \bengalifont ৪ & \bengalifont ৫ & \bengalifont ৬ & \bengalifont ৭ & \bengalifont ৮ & \bengalifont ৯ & 1.4\\

Khmer & \khmerfont ០ & \khmerfont ១ & \khmerfont ២ & \khmerfont ៣ & \khmerfont ៤ & \khmerfont ៥ & \khmerfont ៦ & \khmerfont ៧ & \khmerfont ៨ & \khmerfont ៩ & 0.7\\

Gujarati & \gujaratifont ૦ & \gujaratifont ૧ & \gujaratifont ૨ & \gujaratifont ૩ & \gujaratifont ૪ & \gujaratifont ૫ & \gujaratifont ૬ & \gujaratifont ૭ & \gujaratifont ૮ & \gujaratifont ૯ & 0.5\\

Oriya & \odiyafont \char"0B66 & \odiyafont \char"0B67 & \odiyafont \char"0B68 & \odiyafont \char"0B69 & \odiyafont \char"0B6A & \odiyafont \char"0B6B & \odiyafont \char"0B6C & \odiyafont \char"0B6D & \odiyafont \char"0B6E & \odiyafont \char"0B6F & 0.5\\

Malayalam & \malayalamfont ൦ & \malayalamfont ൧ & \malayalamfont ൨ & \malayalamfont ൩ & \malayalamfont ൪ & \malayalamfont ൫ & \malayalamfont ൬ & \malayalamfont ൭ & \malayalamfont ൮ & \malayalamfont ൯ & 0.4\\

Myanmar & \myanmarfont ၀ & \myanmarfont ၁ & \myanmarfont ၂ & \myanmarfont ၃ & \myanmarfont ၄ & \myanmarfont ၅ & \myanmarfont ၆ & \myanmarfont ၇ & \myanmarfont ၈ & \myanmarfont ၉ & 0.4\\

Telugu & 
\telugufont ౦ & \telugufont ౧ & \telugufont ౨ & \telugufont ౩ & \telugufont ౪ & \telugufont ౫ & \telugufont ౬ & \telugufont ౭ & \telugufont ౮ & \telugufont ౯ & 0.4\\

Thai & \thaifont ๐ & \thaifont ๑ & \thaifont ๒ & \thaifont ๓ & \thaifont ๔ &  \thaifont ๕ & \thaifont ๖ & \thaifont ๗ & \thaifont ๘ & \thaifont ๙ & 0.4\\


Chinese\footnotesize{*} 
    & \begin{otherlanguage}{chinese-simplified} 〇 \end{otherlanguage}
    & \begin{otherlanguage}{chinese-simplified} 一 \end{otherlanguage}
    & \begin{otherlanguage}{chinese-simplified} 二 \end{otherlanguage}
    & \begin{otherlanguage}{chinese-simplified} 三 \end{otherlanguage}
    & \begin{otherlanguage}{chinese-simplified} 四 \end{otherlanguage}
    & \begin{otherlanguage}{chinese-simplified} 五 \end{otherlanguage}
    & \begin{otherlanguage}{chinese-simplified} 六 \end{otherlanguage}
    & \begin{otherlanguage}{chinese-simplified} 七 \end{otherlanguage}
    & \begin{otherlanguage}{chinese-simplified} 八 \end{otherlanguage}
    & \begin{otherlanguage}{chinese-simplified} 九 \end{otherlanguage}
    & 0.3 \\

Kannada & \kannadafont ೦ & \kannadafont ೧ & \kannadafont ೨ & \kannadafont ೩ & \kannadafont ೪ & \kannadafont ೫ & \kannadafont ೬ & \kannadafont ೭ & \kannadafont ೮ & \kannadafont ೯ & 0.2\\

N'ko & \nkofont \char"07C0 & \nkofont \char"07C1  & \nkofont \char"07C2  & \nkofont \char"07C3  & \nkofont \char"07C4  & \nkofont \char"07C5  & \nkofont \char"07C6  & \nkofont \char"07C7  & \nkofont \char"07C8  & \nkofont \char"07C9  & 0.1\\

Tamil & \tamilfont ௦ & \tamilfont ௧ & \tamilfont ௨ & \tamilfont ௩ & \tamilfont ௪ & \tamilfont ௫ & \tamilfont ௬ & \tamilfont ௭ & \tamilfont ௮ & \tamilfont ௯ & 0.1\\

Lao & \laofont ໐ & \laofont ໑ & \laofont ໒ & \laofont ໓ & \laofont ໔ & \laofont ໕ & \laofont ໖ & \laofont ໗ & \laofont ໘ & \laofont ໙ & $2e^{-3}$\\

Ol Chiki & \olfont ᱐ & \olfont ᱑ & \olfont ᱒ & \olfont ᱓ & \olfont ᱔ & \olfont ᱕ & \olfont ᱖ & \olfont ᱗ & \olfont ᱘ & \olfont ᱙ & $6e^{-4}$\\

Adlam & \adlamfont \char"1E950 & \adlamfont \char"1E951 & \adlamfont \char"1E952 & \adlamfont \char"1E953 & \adlamfont \char"1E954 & \adlamfont \char"1E955 & \adlamfont \char"1E956 & \adlamfont \char"1E957 & \adlamfont \char"1E958 & \adlamfont \char"1E959  & $1e^{-4}$\\

Balinese & 
\balinesefont ᭐ & \balinesefont ᭑ & \balinesefont ᭒ & \balinesefont ᭓ & \balinesefont ᭔ & \balinesefont ᭕ & \balinesefont ᭖ & \balinesefont ᭗ & \balinesefont ᭘ & \balinesefont ᭙ & 0\\

Javanese & \javanesefont ꧐ & \javanesefont ꧑ & \javanesefont ꧒ & \javanesefont ꧓ & \javanesefont ꧔ & \javanesefont ꧕ & \javanesefont꧖ & \javanesefont ꧗ & \javanesefont ꧘ & \javanesefont ꧙ & 0\\

Osmanya & \osmanyafont \char"104A0 & \osmanyafont \char"104A1 & \osmanyafont \char"104A2 & \osmanyafont \char"104A3 & \osmanyafont \char"104A4 & \osmanyafont \char"104A5 & \osmanyafont \char"104A6 & \osmanyafont \char"104A7 & \osmanyafont \char"104A8   & \osmanyafont \char"104A9 & 0\\

\bottomrule
\end{tabular}%
}
\caption{Digits 0--9 across the 21 numeral scripts used in this study. The last column reports the average percentage of digit occurrences for each script, detailed in \Cref{sec:script_corpus_dist}. \footnotesize{*\mbox{Simplified} Chinese.}}


\label{tab:script_variants}
\end{table}

\endgroup

\subsection{Format-Variation Numeral Dataset}
\label{sec:format_dataset}

In a separate setting, all numbers in the 336 arithmetic expressions remain in HA digits, but we systematically vary the \textbf{decimal markers} and \textbf{grouping separators}.
We create \textbf{6 formatting conventions} reflecting diverse international styles shown in \Cref{tab:format_variants}.

\begin{table*}[!]
\centering
\resizebox{\textwidth}{!}{%
\begin{tabular}{|c|c|c|c|c|c|c|}
\toprule
\textbf{Format} & \textbf{Decimal Marker} & \textbf{Grouping Separator} & \textbf{Grouping Pattern} & \textbf{Regions}\footnotesize{*} & \textbf{Example} & \textbf{\% Data}\\
\midrule
F1 & Period (.) & Comma (,) & 3-3-3 & North America, United Kingdom, Thailand & 922,436.38 & 63.33\\
F2 & Comma (,) & Period (.) & 3-3-3 & Europe, Latin America & 922.436,38 & 6.61\\
F3 & Comma (,) & Thin space (\textvisiblespace) & 3-3-3 & France, Switzerland & 922\textvisiblespace436,38 & 10.22\\

F4 & Period (.) & Thin space (\textvisiblespace) & 3-3-3 & Russia, Ukraine & 922\textvisiblespace436.38 & 8.25\\

F5 & Comma (,) & Apostrophe (‘) & 3-3-3 & Switzerland, Liechtenstein & 922'436,38 & 2.08\\
F6 & Period (.) & Comma (,) & 3-2-2 & India, Bangladesh & 9,22,436.38 & 9.51\\
\bottomrule
\end{tabular}%
}

\caption{Decimal and grouping separator conventions for six formatting variants. The last column reports the average proportion of count of occurrences for each format, detailed in \Cref{sec:format_corpus_dist}.\footnotesize{*Not a comprehensive list.}}

\label{tab:format_variants}
\end{table*}

\paragraph{} 

Together, these two dataset\footnote{Dataset to be released on Hugging Face post-publication.} variants provide a controlled yet diverse test bed for evaluating how LLMs handle arithmetic when numerical surface forms differ from the HA, Western-centric numeric formatting conventions that dominate pretraining corpora. A detailed description of the dataset generation and annotation process is provided in \Cref{sec:dataset_generation}. We evaluate a total of nine language models, comprising four large and five small models. The complete list of models used in this study, along with details of their configurations and the evaluation protocol, is provided in \Cref{sec:Model_Configuration}.

\section{Multiscript Numerical Reasoning}
\label{sec:numeric_reasoning}

We evaluate multilingual numerical reasoning using the dataset defined in \Cref{sec:script_dataset} across three tasks: (i) identifying the numeral script of a given number, (ii) translating numerals from diverse scripts into standard HA digits, and (iii) performing arithmetic computations directly on numerals expressed in different scripts.

The first two easier tasks serve primarily as probing tasks to assess whether LLMs can recognize and interpret numerals across scripts, while the third task constitutes the main focus of this work, i.e. evaluating models’ ability to perform numerical reasoning under script variation.

\subsection{Script Identification and Translation}
For script identification, we select 30 arithmetic expressions in each of 21 scripts, yielding 630 examples. The expected output is the script name, though we also accept the name of a language that uses that script (e.g., Hindi for Devanagari); verified through manual inspection. For translation, we use a fixed set of 30 unique numerical values rendered in 20 scripts, resulting in 600 test cases, and count translations as correct only if they exactly match the target HA numeral string.

\begin{table}[t]
\centering
\resizebox{\columnwidth}{!}{%
\begin{tabular}{l|cc|cc|cc|cc}
\toprule

\textbf{Script} 
& \multicolumn{2}{c}{\makecell{\textbf{Claude} \\ \textbf{Sonnet 4.5}}}
& \multicolumn{2}{c}{\textbf{GPT-4o}}
& \multicolumn{2}{c}{\makecell{\textbf{Gemini 2.5} \\ \textbf{Pro}}}
& \multicolumn{2}{c}{\makecell{\textbf{Llama-3.3} \\ \textbf{70B}}} \\
 & SI & TR & SI & TR & SI & TR & SI & TR\\

\midrule

Hindu-Arabic & 1.0 & - & 1.0 & - & 1.0 & - & 1.0 & - \\
Adlam & 1.0 & 1.0 & 1.0 & 1.0 & 1.0 & 1.0 & 1.0 & 0.0 \\
Balinese  & 1.0 & 1.0 & 1.0 & 1.0 & 0.97 & 0.97 & 1.0 & 1.0 \\
Bengali & 1.0 & 1.0 & 1.0 & 1.0 & 1.0 & 1.0 & 1.0 & 1.0 \\
Chinese & 1.0 & 1.0 & 0.8 & 0.8 & 0.77 & \textcolor{rust}{\textbf{0.37}} & 0.77 & \textcolor{rust}{\textbf{0.2}} \\
Devanagari & 1.0 & 1.0 & 1.0 & 1.0 & 1.0 & 1.0 & 1.0 & 1.0 \\
Gujarati & 1.0 & 1.0 & 1.0 & 1.0 & 1.0 & 1.0 & 1.0 & 1.0\\
Javanese & 1.0 & 1.0 & \textcolor{rust}{\textbf{0.5}} & \textcolor{rust}{\textbf{0.5}} & 1.0 & 1.0 & 1.0 & 1.0 \\
Kannada & 1.0 & 1.0 & 1.0 & 1.0  & 1.0 & 1.0 & 1.0 & 1.0\\
Khmer & 1.0 & 1.0 & 1.0 & 1.0 & 1.0 & 1.0 & 1.0 & 1.0 \\
Lao  & 1.0  & 1.0 & 0.93 & 0.93 & 0.87 & 0.87 & 1.0 & 1.0 \\
Malayalam & 1.0 & 1.0 & 1.0 & 1.0 & 1.0 & 1.0 & 1.0 & 1.0 \\
Myanmar & 1.0 & 1.0 & 1.0 & 1.0 & 0.77 & \textcolor{rust}{\textbf{0.43}} & 1.0 & 0.0 \\
N'ko & 1.0 & 1.0 & 1.0 & 1.0 & 0.97 & 0.97 & 1.0 & 1.0\\
Ol Chiki  & 1.0 & 1.0  & 1.0 & 1.0 & \textcolor{rust}{\textbf{0.5}} & \textcolor{rust}{\textbf{0.5}} & 1.0 & 1.0 \\
Oriya & 1.0 & 1.0 & 1.0 & 1.0 & 1.0 & 1.0 & 1.0 & 1.0 \\
Osmanya & \textcolor{rust}{\textbf{0.07}} & \textcolor{rust}{\textbf{0.07}} & \textcolor{rust}{\textbf{0.47}} & \textcolor{rust}{\textbf{0.47}} & 0.97 & 0.97 & \textcolor{rust}{\textbf{0.13}} & \textcolor{rust}{\textbf{0.13}} \\
Perso-Arabic & 1.0 & 0.67 & 1.0 & 1.0 & 0.83 & 1.0 & 1.0 & 1.0 \\
Tamil & 1.0 & 1.0 & 1.0 & 1.0 & 0.97 & 1.0 & 1.0 & 1.0 \\
Telugu & 1.0 & 1.0 & 1.0 & 1.0 & 1.0 & 1.0 & 1.0 & 1.0 \\
Thai & 1.0 & 1.0 & 1.0 & 1.0 & 1.0 & 1.0 & 1.0 & 1.0 \\

\midrule
\textbf{Avg. Accuracy} & \textbf{0.96} & \textbf{0.92} & \textbf{0.94} & \textbf{0.89} & \textbf{0.94} & \textbf{0.89} & \textbf{0.95} & \textbf{0.77} \\
\bottomrule
\end{tabular}%
}
\caption{Accuracy on the script identification and numeral translation tasks for LLMs. For each model, the left column (SI) reports accuracy on the script identification task, and the right column (TR) reports accuracy on the numeral translation task. Cells highlighted in \textcolor{rust}{red} indicate cases where model accuracy is 0.5 or lower.}
\label{tab:script_probing_task}
\end{table}

Results for large models, shown in \Cref{tab:script_probing_task}, indicate generally strong performance on both tasks, demonstrating that these models can reliably recognize and translate most numeral scripts. However, consistent failures emerge for very low-resource scripts such as Osmanya, which several models misclassify as Osage. This confusion cannot be attributed to geographic, linguistic, or script-level similarity: Osage is an Indigenous language of North America with a distinct script, whereas Osmanya is used to write Somali in Somalia and parts of East Africa. The fact that multiple models exhibit the same error points to a systematic weakness rather than random noise, likely reflecting shared blind spots in pretraining data or learned representations for extremely rare numeral scripts.

Performance on translation closely mirrors script identification, with a strong correlation of $0.7022$ ($p = 3.95e{-13}$). When a model fails to correctly identify a numeral's script, it typically also fails to produce an accurate HA translation. This alignment suggests that most translation errors arise from upstream difficulties in script recognition, rather than from an inability to convert numeric values once the script is correctly identified.

We also considered a number of smaller, open source language models. 
Results presented in 
\Cref{sec:multilingual_small_llm} show most smaller language models achieve near-zero accuracy on both tasks for most scripts, with a few exceptions for higher-resource scripts such as Devanagari and Perso-Arabic.
Due to this consistently poor performance, we focus the rest of our analysis on the four larger models.

\subsection{Arithmetic Computation}
We evaluate each model's ability to perform basic arithmetic operations, i.e. addition, subtraction, multiplication and division, when the numerals in the expression are presented in HA and non–HA scripts, using the dataset from \Cref{sec:script_dataset}.

\subsubsection{Impact of Numeral Script Variation on Arithmetic Performance}
To isolate the effect of numeral script changes on model performance, we only alter the digits in the operands while keeping the arithmetic term and prompt in English. An example prompt is shown in \Cref{sec:multilingual_prompt_examples}.

\begin{figure}[!]
    \centering
    \includegraphics[width=1\linewidth]{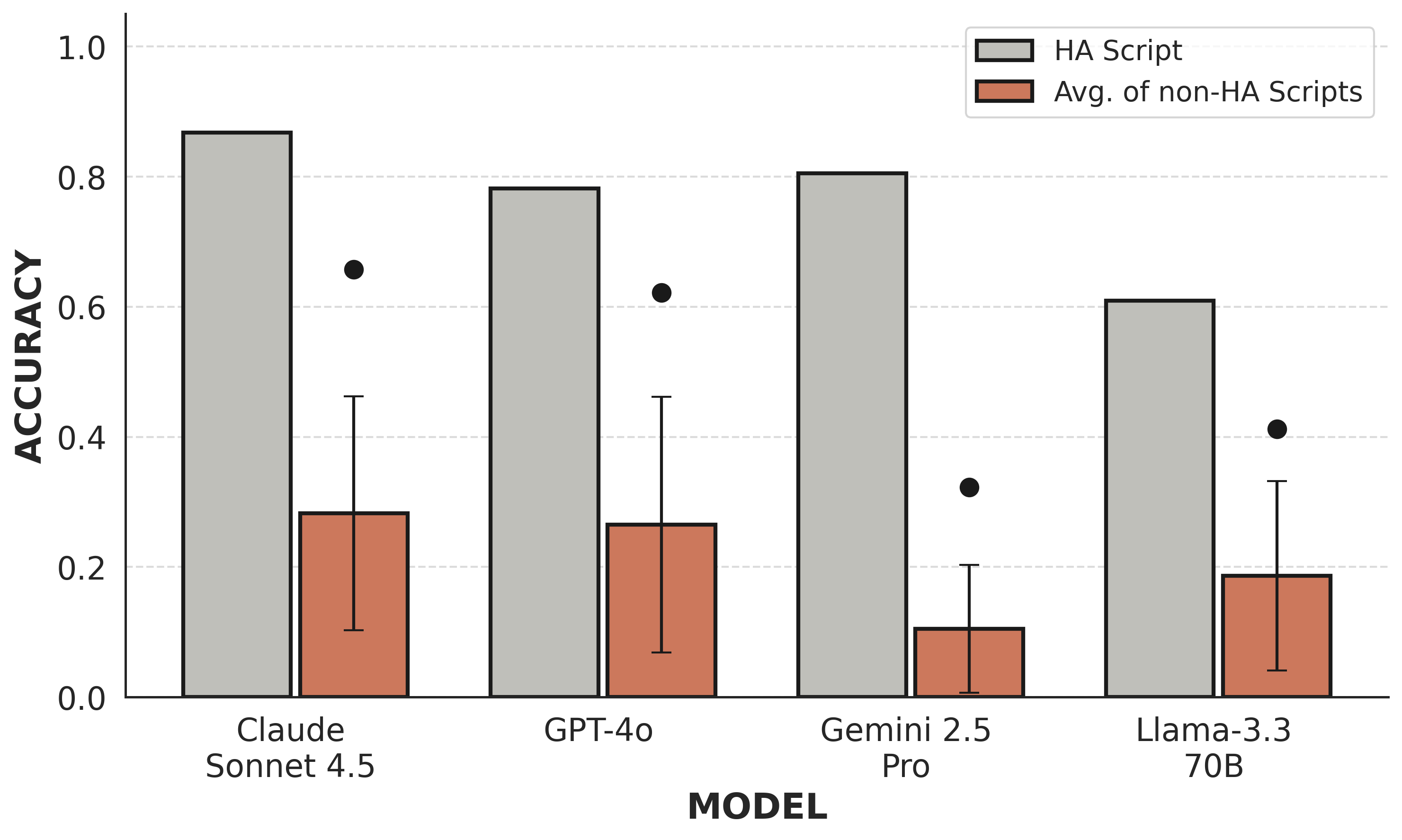}
    \caption{Comparison of model accuracy on the arithmetic task. Error bars represent the standard deviation and the dots represent the maximum accuracy among non-HA scripts.}
    \label{fig:org_v_all_script_accuracy_bar}
\end{figure}

As shown in \Cref{fig:org_v_all_script_accuracy_bar}, shifting numerals from the standard HA script to other non-HA scripts results in an average performance drop of approximately 66--87\%, with Gemini exhibiting the lowest performance among the evaluated models.

\subsubsection{Bridging the Performance Gap through Prompting Strategies}

\begin{figure*}[!]
    \centering
    \includegraphics[width=0.8\textwidth]{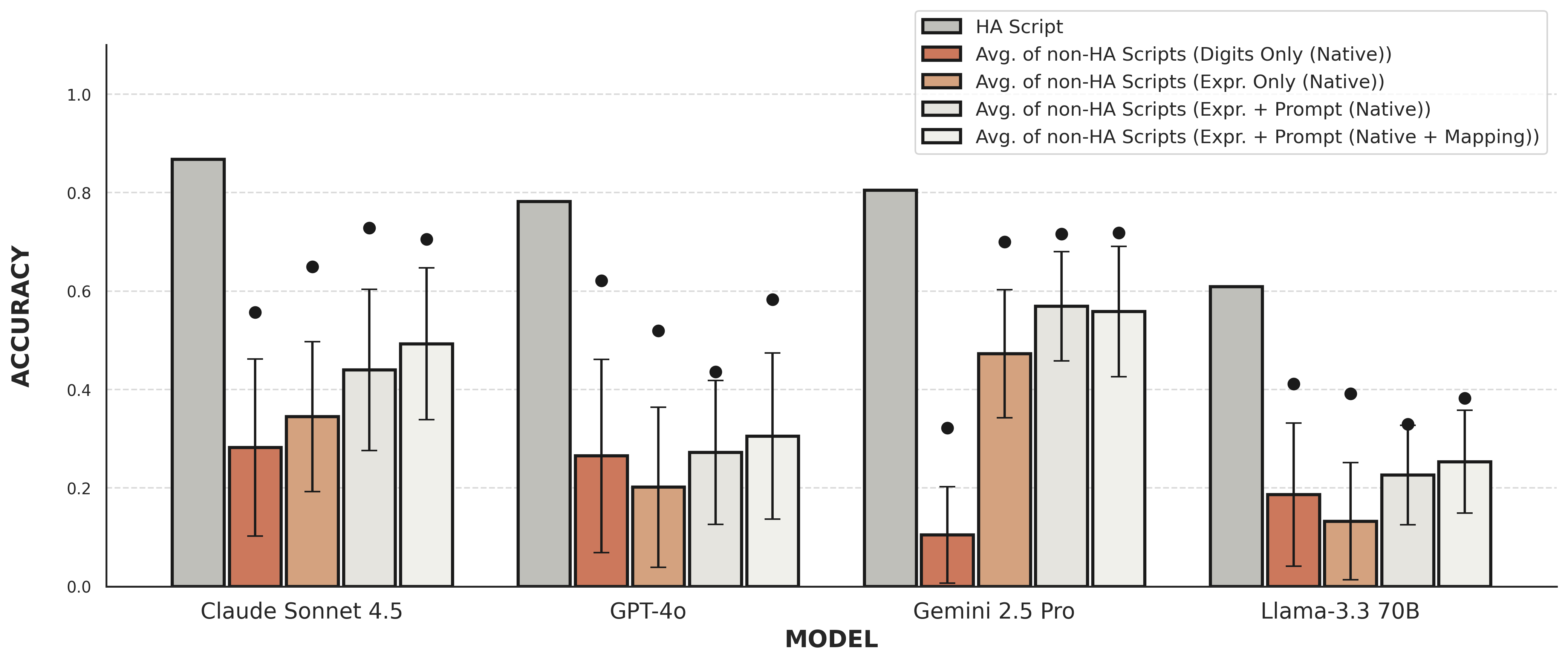}
    \caption{Comparison of model accuracy on the arithmetic task. Error bars represent the standard
deviation and the dots represent the maximum accuracy among non-HA scripts}
    \label{fig:org_v_all_script_accuracy_all_prompts_bar}
\end{figure*}

To improve model performance on non-HA scripts, we experimented with various prompt modifications. First, we provided additional hints, such as explicitly mentioning the name of the script used. However, this approach did not yield any noticeable improvement. Next, we translated the operator word into the script corresponding to the numerals. For the words used in the prompts, we selected one language that uses the same script as the numbers. The complete script to language mapping used in this study is provided in \Cref{tab:script_language_mapping} in \Cref{sec:script_arithmetic_computation}. This strategy, as shown in \Cref{fig:org_v_all_script_accuracy_all_prompts_bar}, results in a substantial performance improvement compared to using the operator in English.

We further observe in \Cref{fig:org_v_all_script_accuracy_all_prompts_bar} that presenting the entire user prompt in the native language using the script in question yields strong performance across the models, highlighting the importance of consistent script usage and minimal cross-script interference during arithmetic reasoning. The largest performance improvement, however, is observed when models are provided with an explicit mapping between HA and non-HA digits. All prompts used for this task are provided in \Cref{sec:multilingual_prompt_examples}.

\subsection{Quantifying Factors Affecting Multiscript Arithmetic Performance}
\label{sec:math_script}










\begin{table}[!]
\centering
\setlength{\extrarowheight}{1.75pt}
\resizebox{\linewidth}{!}{%
\begin{tabular}{l|c|c}

\toprule
\textbf{Feature} &  \textbf{Coefficient} & \textbf{$p$-Value} \\
\midrule

Intercept & 1.276 & 0.0965 \\
Total \# Operand Digits & -0.027 & 0.098 \\
Subtraction & -0.082 & 0.778 \\
Multiplication & -1.136 & 0.022 \\
Division & -0.035 & 0.904 \\
Tokens per Digit & -0.198 & $<1e^{-8}$\\

\hline

Bengali & -2.189 & $<1e^{-8}$ \\
Chinese & -3.544 & $<1e^{-8}$ \\
Devanagari & -1.995 & $<1e^{-8}$ \\
Gujarati & -2.692 & $<1e^{-8}$ \\
Kannada & -2.02 & $<1e^{-8}$ \\
Khmer & -2.01 & $<1e^{-8}$ \\
Lao & -2.747 & $<1e^{-8}$ \\
Malayalam & -3.118 & $<1e^{-8}$ \\
Myanmar & -2.335 & $<1e^{-8}$ \\
Ol Chiki & -3.09 & $<1e^{-8}$ \\
Oriya & -2.63 & $<1e^{-8}$ \\
Perso-Arabic & -1.287 & $<1e^{-8}$ \\
Tamil & -3.406 & $<1e^{-8}$ \\
Telugu & -2.794 & $<1e^{-8}$ \\
Thai & -2.233 & $<1e^{-8}$ \\

\hline

Expr. Only (Native) & 0.369 & $<1e^{-8}$ \\
Expr. + Prompt (Native) & 0.238 & $<1e^{-8}$ \\
Expr. + Prompt (Native + Mapping) & 0.581 & $<1e^{-8}$ \\

\bottomrule
\end{tabular}%
}
\caption{Coefficient estimates and $p$-value from a GLMER model.}

\label{tab:mixed_effects_scripts}
\end{table}

To assess the factors underlying model performance on arithmetic reasoning across diverse numeral systems, we employ a logistic generalized linear mixed-effects model (GLMER; \citealp{Breslow01031993}; \citealp{lme4}). A mixed-effects formulation is necessary because each arithmetic expression appears multiple times across different numeral scripts and prompting conditions, violating the independence assumption required by logistic regression \cite{winter2013linearmodelslinearmixed}. By modeling both fixed effects and random effects, the GLMER allows us to account for repeated measurements over the same problems and models while isolating the influence of representational factors. The model is specified as:
\begin{align*}
    \textsf{response} \sim{} 
        & \textsf{total\_digits} + \textsf{operation} + \textsf{script} \\
        & + \textsf{tokens\_per\_digit} + \textsf{prompt} \\
        & + (1 \mid \textsf{model}) + (1 \mid \textsf{index})
\end{align*}
The fixed effects include the total number of digits in the operands, the arithmetic operation, the number of tokens per digit, the numeral script, and the prompting strategy. Random intercept effects are specified for the language model and the problem index. The factor reference levels are the HA numeral system, the addition operator, and the English operator prompt. These levels have a coefficient of 0. The model coefficient estimates and $p$-values are shown in \Cref{tab:mixed_effects_scripts}.

Gemini~2.5~Pro was excluded from this analysis due to its consistently low accuracy, which created an extreme imbalance in the response variable making it very difficult to estimate these regression coefficients. Likewise, the numeral scripts Adlam, Osmanya, N’Ko, Balinese, and Javanese were omitted because model performance on these scripts was often zero, making it impossible to reliably estimate their effects.

\paragraph{Influence of Operation and Tokenization}
Among the arithmetic operations, \textit{multiplication} exhibits a significant negative effect relative to addition. This is consistent with the increased computational complexity of multiplication, particularly when operands contain a larger number of digits, resulting in longer outputs.

The \textit{tokens per digit} feature shows a strong and statistically significant negative coefficient which confirms that scripts that are tokenized more efficiently---requiring fewer subword units---tend to be processed more accurately. 

\paragraph{Impact of Numeral Scripts}
A key finding of our analysis is the consistent negative impact of non–HA numeral scripts on arithmetic accuracy. All scripts included in the model yield statistically significant negative coefficients relative to the HA reference level. This pattern reveals what we refer to as a \textit{script tax}: a systematic performance penalty incurred when models process arithmetic expressed in non–HA scripts. While LLMs possess robust arithmetic capabilities, these abilities appear to be strongly tied to the numeral representations most prevalent in their pretraining corpora.

\paragraph{Prompt Engineering Effects}
Prompting strategies play a significant role in mitigating performance degradation. The \textit{Expr. + Prompt (Native + Mapping)} strategy provides the largest positive effect ($0.581$), suggesting that explicitly mapping native-script numerals to a familiar reference representation partially alleviates the challenges posed by non-standard scripts. This result highlights the effectiveness of prompt-level interventions in bridging representation gaps without modifying model parameters.

\section{Arithmetic Reasoning with Formatted Numbers}
\label{sec:arithmetic_reasoning}

In the United States, among other regions, numbers typically use a period as the decimal separator and a comma as the thousands marker, grouping digits into sets of three. As shown in \Cref{tab:format_variants}, this convention (F1) appears frequently in large pretraining corpora and we believe it is therefore familiar to most LLMs. In contrast, other formatting styles—such as using an apostrophe as the grouping separator or grouping digits into pairs after the first thousand—appear far less often and are likely underrepresented during training.

\begingroup
\setlength{\tabcolsep}{2pt} 
\renewcommand{\arraystretch}{0.9} 

\begin{table*}[]
\centering
\small
\setlength{\extrarowheight}{10pt}
\resizebox{\textwidth}{!}{%
\begin{tabular}{|c|l|c|c|c|}
\toprule

\textbf{Format} & \textbf{Sample} & \makecell{\textbf{Claude} \\ \textbf{Sonnet 4.5}} & \textbf{GPT 4o} & \textbf{Gemini~2.5~Pro}\\
\midrule

F1 & \ttfamily{922,436.38 + 4,359} & 12 
& \token{922}\token{,}\token{436}\token{.}\token{38}\token{ +}\token{ }\token{4}\token{,}\token{359} 
& \token{9}\token{2}\token{2}\token{,}\token{4}\token{3}\token{6}\token{.}\token{3}\token{8}\token{ +}\token{ }\token{4}\token{,}\token{3}\token{5}\token{9} \\

F2 & \ttfamily{922.436,38 + 4.359} & 12 
& \token{922}\token{.}\token{436}\token{,}\token{38}\token{ +}\token{ }\token{4}\token{.}\token{359} 
& \token{9}\token{2}\token{2}\token{.}\token{4}\token{3}\token{6}\token{,}\token{3}\token{8}\token{ +}\token{ }\token{4}\token{.}\token{3}\token{5}\token{9} \\

F3 & \ttfamily{922\textvisiblespace436,38 + 4\textvisiblespace359} & 12 
& \token{922}\token{\textvisiblespace}\token{436}\token{,}\token{38}\token{ +}\token{ }\token{4}\token{\textvisiblespace}\token{359} &
\token{9}\token{2}\token{2}\unknowntokene{$^1$}\unknowntokene{$^2$}\unknowntokene{$^3$}\token{4}\token{3}\token{6}\token{,}\token{3}\token{8}\token{ +}\token{ }\token{4}\unknowntokene{$^1$}\unknowntokene{$^2$}\unknowntokene{$^3$}\token{3}\token{5}\token{9} \\

F4 & \ttfamily{922\textvisiblespace436.38 + 4\textvisiblespace359} & 12 
& \token{922}\token{\textvisiblespace}\token{436}\token{.}\token{38}\token{ +}\token{ }\token{4}\token{\textvisiblespace}\token{359} 
& \token{9}\token{2}\token{2}\unknowntokene{$^1$}\unknowntokene{$^2$}\unknowntokene{$^3$}\token{4}\token{3}\token{6}\token{.}\token{3}\token{8}\token{ +}\token{ }\token{4}\unknowntokene{$^1$}\unknowntokene{$^2$}\unknowntokene{$^3$}\token{3}\token{5}\token{9} \\

F5 & \ttfamily{922'436,38 + 4'359} & 12 
& \token{922}\token{'}\token{436}\token{,}\token{38}\token{ +}\token{ }\token{4}\token{'}\token{359} 
& \token{9}\token{2}\token{2}\token{'}\token{4}\token{3}\token{6}\token{,}\token{3}\token{8}\token{ +}\token{ }\token{4}\token{'}\token{3}\token{5}\token{9} \\

F6 & \ttfamily{9,22,436.38 + 4,359} & 14
& \token{9}\token{,}\token{22}\token{,}\token{436}\token{.}\token{38}\token{ +}\token{ }\token{4}\token{,}\token{359} 
& \token{9}\token{,}\token{2}\token{2}\token{,}\token{4}\token{3}\token{6}\token{.}\token{3}\token{8}\token{ +}\token{ }\token{4}\token{,}\token{3}\token{5}\token{9} \\

\bottomrule
\end{tabular}
}
\caption{Tokenization of input equations across the six formatting variants for various LLMs (The tokenization for Llama~3.3 is the same as that of GPT 4o). For Claude Sonnet~4.5, the column shows only the number of tokens, as the tokenizer is not publicly accessible. Gemini 2.5 Pro splits the thin space into 3 single-byte tokens.}
\label{tab:format_tokenization}
\end{table*}

\endgroup

We evaluate each model’s ability to perform basic arithmetic on a single arithmetic task when the same HA numerals are presented using \textbf{different formatting conventions}, varying both the \textit{decimal marker} and the \textit{digit-grouping separator}, using the dataset defined in \Cref{sec:format_dataset}. By holding the script constant and altering only the formatting, this evaluation isolates the effect of typographic variation, allowing us to assess how robust LLMs are to differences in number presentation that occur across various writing systems.

\subsection{The Baseline}
The baseline arithmetic task uses unformatted input expressions, with the model as expected producing unformatted outputs. For the six formatted versions, the formatted expressions are provided as input to the language models, and outputs in any format (or unformatted) are accepted. (An example prompt is provided in \Cref{sec:format_prompt_examples}.)

\begin{figure}[!]
    \centering
    \includegraphics[width=1\linewidth]{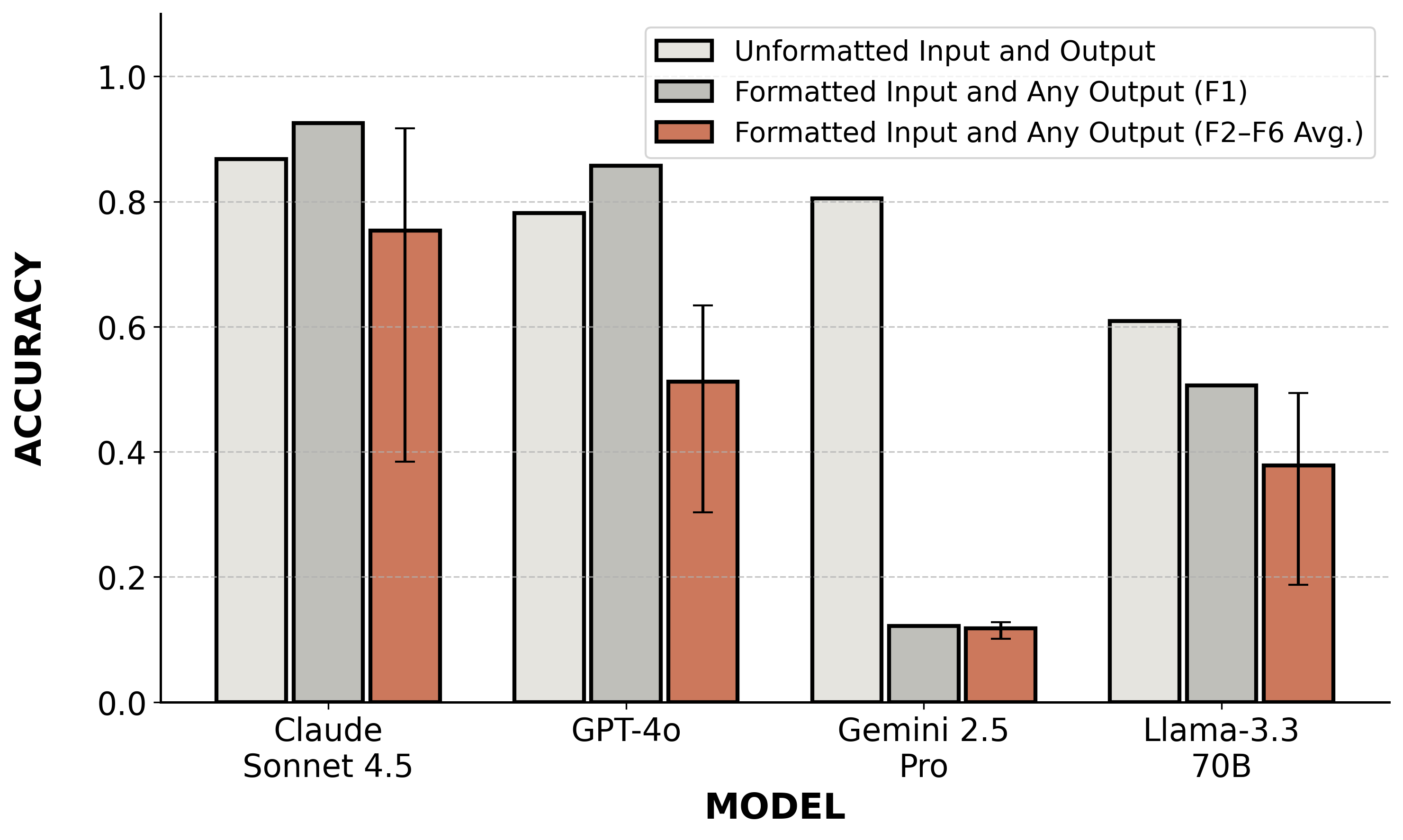}
    \caption{Comparison of model accuracy on the arithmetic task across formatted variants. Error bars represent the standard deviation.}
    \label{fig:baseline_format_bar}
\end{figure}

As shown in \Cref{fig:baseline_format_bar}, for some models, performance on the F1 format is comparable to or better than performance on the baseline arithmetic expressions, which is consistent with prior findings ~\cite{singh2024tokenizationcountsimpacttokenization} that adding commas correctly helps. While accuracy on formats F2--F6 is generally lower, the results suggest that models are largely able to treat formatted numbers in arithmetic expressions the same as their unformatted representations. One notable exception is Gemini~2.5~Pro, which exhibits consistently low performance when arithmetic expressions are presented with formatting. While differences in model performance may stem from several factors---such as model size, pretraining data composition, or fine-tuning strategies---these details are not publicly available for many large models, particularly closed-source ones.

To better understand the observed performance differences across models, we analyze the tokenization strategies employed in this study. As shown in \Cref{tab:format_tokenization}, Gemini-2.5 Pro is the only model that tokenizes each digit as an individual token. In contrast, the other models exhibit broadly similar subword segmentation patterns, typically segmenting into sequences of up to 3 digits.

While digit-level tokenization alone cannot fully account for Gemini~2.5~Pro’s lower performance, it aligns with prior work \cite{kreitner2025efficientnumeracylanguagemodels} showing that highly fragmented numeral representations can hinder numerical reasoning. Taken together, these results suggest that tokenization strategy is an important contributing factor, but one that likely interacts with other aspects of model training and architecture rather than acting in isolation.

\subsection{Prompting Strategies}

To evaluate whether models can reproduce the typographic conventions present in the input numbers, we prompt them to return the answer in the same format as the numbers in the input arithmetic expressions. We experiment with the following prompting strategies. Example prompts for each strategy are provided in the \Cref{sec:format_prompt_examples}.

\begin{figure*}[h]
    \centering
    \includegraphics[width=0.9\textwidth]{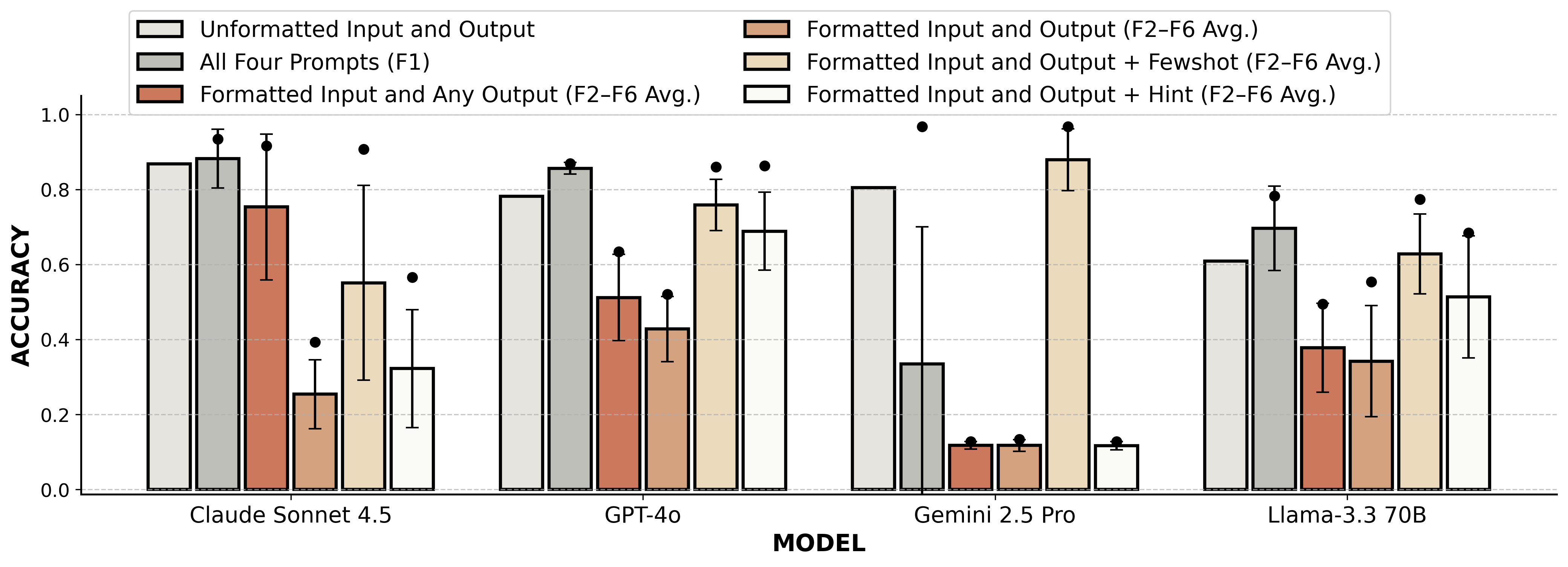}
    \caption{Comparison of model accuracy on the arithmetic task. Error bars denote standard deviation, and dots indicate the maximum accuracy across formats. For F1, the bar shows the average over the four prompting strategies, while the dot represents the best-performing prompt.}
    \label{fig:format_all_prompts}
\end{figure*}

\textbf{Formatted Input and Output:} The model is expected to return the answer in the same format as the input, without any additional guidance. Note that this is a more difficult problem than the Baseline which accepted the answer in any format. 

\textbf{Formatted Input and Output + Hint:} The prompt also explicitly specifies the decimal marker and grouping separator used in the input arithmetic expressions.

\textbf{Formatted Input and Output + Fewshot:} The prompt also includes two examples of formatted input arithmetic expressions along with their corresponding outputs.

As shown in \Cref{fig:format_all_prompts}, performance is highest when a period is used as the decimal marker and a comma is used as the grouping separator, with digits grouped in threes (F1). In contrast, model performance is lowest when these symbols are reversed (F2).\footnote{A more detailed error analysis of LLM performance across the individual formatting variants and prompting strategies is provided in \Cref{sec:error_analysis}.} While this could partly be attributed to the relative scarcity of such formats in training data, further analysis reveals an interesting pattern. A substantial portion of training examples whose digit-grouping pattern matches format F2 (i.e. period-separated digit groups) appear in contexts such as IP addresses rather than arithmetic expressions. As a result, models are frequently exposed to these patterns without any associated numerical computation context, which likely leads to confusion during arithmetic reasoning.

The format with the second-lowest performance, F3, uses a thin space as the grouping separator with a comma as the decimal marker, a convention common in France and Switzerland. This demonstrates that even purely typographic deviations from familiar grouping patterns, when underrepresented in pretraining, can significantly impact arithmetic accuracy.

Prompting strategies can mitigate these issues to some extent. Few-shot prompting with numeric examples increases model performance by 20--60\%, demonstrating that models can adapt through in-context learning. This also suggests that much of the failure in zero-shot scenarios is not due to inability to compute but rather due to a lack of context for reproducing the formatting.

\subsection{Quantifying the Impact of Numeric Formatting on Arithmetic Performance}

\begin{table}[!h]
\centering
\setlength{\extrarowheight}{1.75pt}
\resizebox{\linewidth}{!}{%
\begin{tabular}{l|c|c}

\toprule
\textbf{Feature} &  \textbf{Coefficient} & \textbf{$p$-Value} \\
\midrule

Intercept & 3.349 & $<1e^{-8}$ \\
Total \# Operand Digits & -0.154 & $<1e^{-8}$ \\
Subtraction & -0.230 & 0.196 \\
Multiplication & -2.951 & $<1e^{-8}$ \\
Division & -0.083 & $<1e^{-8}$ \\
Tokens per Digit & -0.514 & $<1e^{-8}$\\
\hline

F2 & -2.259 & $<1e^{-8}$ \\
F3 & -1.460 & $<1e^{-8}$ \\
F4 & -0.991 & $<1e^{-8}$ \\
F5 & -1.278 & $<1e^{-8}$ \\
F6 & -1.342 & $<1e^{-8}$ \\
\hline

Fmt. Input → Fmt. Output & -0.793 & $<1e^{-8}$ \\
Fmt. Input + Hint → Fmt. Output & -0.075 & 0.049 \\
Fmt. Input + Few-shot → Fmt. Output & 1.694 & $<1e^{-8}$ \\

\bottomrule
\end{tabular}%
}
\caption{Coefficient estimates and $p$-value from a GLMER model.}

\label{tab:mixed_effects_format}
\end{table}

To analyze the factors affecting arithmetic accuracy under different numeric formatting conventions, we fit a GLMER model analogous to that described in \Cref{sec:math_script}, replacing the \textsf{script} factor with \textsf{format} and using different prompting strategies in \textsf{prompt}. The reference levels are the F1 formatting convention and the addition operator.



The model coefficients shown in \Cref{tab:mixed_effects_format}, reveal a strong and consistent impact of numeric formatting on arithmetic performance. All alternative formatting variants (F2–F6) exhibit large negative coefficients relative to F1, indicating that arithmetic expressions formatted differently from the standard F1 convention are substantially more difficult for LLMs to solve. The magnitude of these effects shows that even purely typographic changes that do not alter the underlying numerical values can significantly impair model performance.

Beyond formatting, the type of arithmetic operation also influences accuracy. As with the previous model, multiplication is significantly more challenging than addition, while subtraction and division show smaller or less consistent effects. As expected, increasing the total number of operand digits shows reduced accuracy, reflecting the greater complexity of longer numerical expressions.

Prompting strategies partially mitigate these challenges. Requiring models to produce formatted outputs further degrades performance, whereas explicitly providing formatting hints yields only negligible effects. In contrast, few-shot prompting produces a substantial positive effect ($1.694$, $p < 1e^{-8}$), suggesting that exposure to representative examples enables models to better adapt to unfamiliar numeric conventions.

\section{Related Work}
Research on numerical reasoning in LLMs has examined arithmetic competence from several complementary perspectives, including representation choices, internal computation strategies, and multilingual generalization.

A growing line of work investigates how representational conventions influence numerical reasoning. \citet{bui-etal-2025-generalization} study the role of measurement systems and show that LLMs perform best when outputs align with the cultural and contextual norms implied by the prompt (e.g., USD for U.S.-centric questions, kilometers for European contexts). Their findings suggest that reasoning is tightly coupled with learned contextual expectations, which aligns with our observation that arithmetic performance improves when numerals are presented in scripts and formats that are common in the data. Similarly, \citet{kreitner2025efficientnumeracylanguagemodels} survey numeracy across frontier models and propose a single token number format aimed at improving efficiency and consistency, highlighting the importance of how numeric information is encoded.

Several studies focus on the internal mechanisms LLMs use to perform arithmetic. \citet{baeumel-etal-2025-lookahead} analyze addition behavior and find that models rely on a limited single-digit lookahead heuristic when handling carries, exposing structural limitations in multi-digit arithmetic. \citet{jobanputra2025llmssubtractnumbers} provide a detailed study of subtraction, showing that errors increase substantially when the result is negative.

Multilingual numerical reasoning has primarily been studied in the context of math word problems. The MGSM benchmark \citep{shi2022languagemodelsmultilingualchainofthought}, a multilingual extension of GSM8K, evaluates math reasoning across languages but retains Hindu–Arabic digits, thereby isolating linguistic variation while holding numeric representation constant. Follow-up work by \citet{peter2025mindgapnottranslation} revisits MGSM and attributes much of the observed performance gap to translation artifacts and answer extraction issues rather than intrinsic mathematical deficiencies. In a related direction, \citet{bhattacharya-etal-2025-investigating} study multilingual numerical puzzles using number words instead of digits, examining how lexicalized numerals affect reasoning across languages. They also show that models fail to reliably infer the implicit compositional structure of numeral systems that humans readily exploit.

\section{Conclusion}
In this work, we systematically examined how variations in numeral scripts and formatting conventions affect arithmetic performance in large language models. Our results reveal a strong and statistically significant impact of both script and format on LLM numeracy, even when the underlying arithmetic remains unchanged. Deviations from the default Hindu–Arabic, Western-formatted representation impose a substantial script tax, leading to pronounced performance degradation across models. Importantly, we show that this loss is not inevitable: prompting strategies play a critical role in mitigating these effects. In particular, few-shot prompting with explicit numeric examples yields large and statistically significant gains, enabling models to better adapt to unfamiliar scripts and formatting conventions. Prompting with explicit hints has been observed to provide no noticeable improvement. These findings suggest that many observed failures stem less from an inability to perform arithmetic than from a lack of contextual grounding for interpreting non-standard numeric surface forms. Taken together, our results underscore the fragility of numerical reasoning under representational shifts and highlight practical, actionable prompting techniques as an effective way to improve robustness when working with diverse numeral scripts and typographic formats.

\section*{Limitations}
This study focuses on isolating the effects of numeral scripts, formatting, and lightweight prompting strategies on arithmetic performance, and therefore does not exhaustively explore all possible methods for improving model accuracy, such as explicit chain-of-thought prompting or tool-calling. Our goal is to assess robustness under minimal intervention and to avoid introducing additional reasoning scaffolds that may obscure the impact of representational variation itself. Additionally, for the F3 and F4 formatting variants, we restrict digit grouping to a single representative thin-space separator rather than evaluating all typographically acceptable spacing alternatives. This design choice enables controlled comparisons across formats while keeping the evaluation set tractable, but may under-represent the full range of real-world numeric typography. That being said, we do not see any potential risks associated with this work, as the study solely aims to highlight the under-representation of many scripts and numeric formats to inform future improvements.

\bibliography{custom,anthology-1}

\begin{thebibliography}{20}
\providecommand{\natexlab}[1]{#1}

\bibitem[{Baeumel et~al.(2025)Baeumel, Genabith, and Ostermann}]{baeumel-etal-2025-lookahead}
Tanja Baeumel, Josef~Van Genabith, and Simon Ostermann. 2025.
\newblock \href {https://doi.org/10.18653/v1/2025.blackboxnlp-1.15} {The lookahead limitation: Why multi-operand addition is hard for {LLM}s}.
\newblock In \emph{Proceedings of the 8th BlackboxNLP Workshop: Analyzing and Interpreting Neural Networks for NLP}, pages 250--262, Suzhou, China. Association for Computational Linguistics.

\bibitem[{Bates et~al.(2015)Bates, M{\"a}chler, Bolker, and Walker}]{lme4}
Douglas Bates, Martin M{\"a}chler, Ben Bolker, and Steve Walker. 2015.
\newblock \href {https://doi.org/10.18637/jss.v067.i01} {Fitting linear mixed-effects models using {lme4}}.
\newblock \emph{Journal of Statistical Software}, 67(1):1--48.

\bibitem[{Bhattacharya et~al.(2025)Bhattacharya, Papadimitriou, Davidson, and Alvarez-Melis}]{bhattacharya-etal-2025-investigating}
Antara~Raaghavi Bhattacharya, Isabel Papadimitriou, Kathryn Davidson, and David Alvarez-Melis. 2025.
\newblock \href {https://doi.org/10.18653/v1/2025.emnlp-main.1438} {Investigating the interaction of linguistic and mathematical reasoning in language models using multilingual number puzzles}.
\newblock In \emph{Proceedings of the 2025 Conference on Empirical Methods in Natural Language Processing}, pages 28310--28320, Suzhou, China. Association for Computational Linguistics.

\bibitem[{Breslow and Clayton(1993)}]{Breslow01031993}
N.~E. Breslow and D.~G. Clayton. 1993.
\newblock \href {https://doi.org/10.1080/01621459.1993.10594284} {Approximate inference in generalized linear mixed models}.
\newblock \emph{Journal of the American Statistical Association}, 88(421):9--25.

\bibitem[{Bui et~al.(2025)Bui, Park, Glava{\v{s}}, Schmidt, and Wense}]{bui-etal-2025-generalization}
Minh~Duc Bui, Kyung~Eun Park, Goran Glava{\v{s}}, Fabian~David Schmidt, and Katharina Von~Der Wense. 2025.
\newblock \href {https://doi.org/10.18653/v1/2025.acl-long.1032} {On generalization across measurement systems: {LLM}s entail more test-time compute for underrepresented cultures}.
\newblock In \emph{Proceedings of the 63rd Annual Meeting of the Association for Computational Linguistics (Volume 1: Long Papers)}, pages 21262--21276, Vienna, Austria. Association for Computational Linguistics.

\bibitem[{Gao et~al.(2020)Gao, Biderman, Black, Golding, Hoppe, Foster, Phang, He, Thite, Nabeshima, Presser, and Leahy}]{gao2020pile800gbdatasetdiverse}
Leo Gao, Stella Biderman, Sid Black, Laurence Golding, Travis Hoppe, Charles Foster, Jason Phang, Horace He, Anish Thite, Noa Nabeshima, Shawn Presser, and Connor Leahy. 2020.
\newblock \href {https://arxiv.org/abs/2101.00027} {The pile: An 800gb dataset of diverse text for language modeling}.
\newblock \emph{Preprint}, arXiv:2101.00027.

\bibitem[{Jobanputra et~al.(2025)Jobanputra, Walter, Mehta, Veseli, Chapple, Wang, Chetani, Pavlick, Vergari, and Demberg}]{jobanputra2025llmssubtractnumbers}
Mayank Jobanputra, Nils~Philipp Walter, Maitrey Mehta, Blerta Veseli, Evan Parker~Kelly Chapple, Yifan Wang, Sneha Chetani, Ellie Pavlick, Antonio Vergari, and Vera Demberg. 2025.
\newblock \href {https://arxiv.org/abs/2511.02795} {Can llms subtract numbers?}
\newblock \emph{Preprint}, arXiv:2511.02795.

\bibitem[{Kreitner et~al.(2025)Kreitner, Hager, Mengedoht, Kaissis, Rueckert, and Menten}]{kreitner2025efficientnumeracylanguagemodels}
Linus Kreitner, Paul Hager, Jonathan Mengedoht, Georgios Kaissis, Daniel Rueckert, and Martin~J. Menten. 2025.
\newblock \href {https://arxiv.org/abs/2510.06824} {Efficient numeracy in language models through single-token number embeddings}.
\newblock \emph{Preprint}, arXiv:2510.06824.

\bibitem[{McCoy et~al.(2019)McCoy, Pavlick, and Linzen}]{mccoy-etal-2019-right}
R.~Thomas McCoy, Ellie Pavlick, and Tal Linzen. 2019.
\newblock \href {https://doi.org/10.18653/v1/P19-1334} {Right for the wrong reasons: Diagnosing syntactic heuristics in natural language inference}.
\newblock In \emph{Proceedings of the 57th Annual Meeting of the Association for Computational Linguistics}, pages 3428--3448, Florence, Italy. Association for Computational Linguistics.

\bibitem[{Nogueira et~al.(2021)Nogueira, Jiang, and Lin}]{nogueira2021investigatinglimitationstransformerssimple}
Rodrigo Nogueira, Zhiying Jiang, and Jimmy Lin. 2021.
\newblock \href {https://arxiv.org/abs/2102.13019} {Investigating the limitations of transformers with simple arithmetic tasks}.
\newblock \emph{Preprint}, arXiv:2102.13019.

\bibitem[{Ortiz~Su{\'a}rez et~al.(2019)Ortiz~Su{\'a}rez, Sagot, and Romary}]{ortizsuarez}
Pedro~Javier Ortiz~Su{\'a}rez, Beno{\^i}t Sagot, and Laurent Romary. 2019.
\newblock \href {https://doi.org/10.14618/IDS-PUB-9021} {{Asynchronous Pipeline for Processing Huge Corpora on Medium to Low Resource Infrastructures}}.

\bibitem[{Penedo et~al.(2025)Penedo, Kydlíček, Sabolčec, Messmer, Foroutan, Kargaran, Raffel, Jaggi, Werra, and Wolf}]{penedo2025fineweb2pipelinescale}
Guilherme Penedo, Hynek Kydlíček, Vinko Sabolčec, Bettina Messmer, Negar Foroutan, Amir~Hossein Kargaran, Colin Raffel, Martin Jaggi, Leandro~Von Werra, and Thomas Wolf. 2025.
\newblock \href {https://arxiv.org/abs/2506.20920} {Fineweb2: One pipeline to scale them all -- adapting pre-training data processing to every language}.
\newblock \emph{Preprint}, arXiv:2506.20920.

\bibitem[{Peter et~al.(2025)Peter, Vilar, Domhan, Malkin, and Freitag}]{peter2025mindgapnottranslation}
Jan-Thorsten Peter, David Vilar, Tobias Domhan, Dan Malkin, and Markus Freitag. 2025.
\newblock \href {https://arxiv.org/abs/2511.05162} {Mind the gap... or not? how translation errors and evaluation details skew multilingual results}.
\newblock \emph{Preprint}, arXiv:2511.05162.

\bibitem[{Shi et~al.(2022)Shi, Suzgun, Freitag, Wang, Srivats, Vosoughi, Chung, Tay, Ruder, Zhou, Das, and Wei}]{shi2022languagemodelsmultilingualchainofthought}
Freda Shi, Mirac Suzgun, Markus Freitag, Xuezhi Wang, Suraj Srivats, Soroush Vosoughi, Hyung~Won Chung, Yi~Tay, Sebastian Ruder, Denny Zhou, Dipanjan Das, and Jason Wei. 2022.
\newblock \href {https://arxiv.org/abs/2210.03057} {Language models are multilingual chain-of-thought reasoners}.
\newblock \emph{Preprint}, arXiv:2210.03057.

\bibitem[{Singh and Strouse(2024)}]{singh2024tokenizationcountsimpacttokenization}
Aaditya~K. Singh and DJ~Strouse. 2024.
\newblock \href {https://arxiv.org/abs/2402.14903} {Tokenization counts: the impact of tokenization on arithmetic in frontier llms}.
\newblock \emph{Preprint}, arXiv:2402.14903.

\bibitem[{Singha et~al.(2023)Singha, Cambronero, Gulwani, Le, and Parnin}]{singha2023tabularrepresentationnoisyoperators}
Ananya Singha, José Cambronero, Sumit Gulwani, Vu~Le, and Chris Parnin. 2023.
\newblock \href {https://arxiv.org/abs/2310.10358} {Tabular representation, noisy operators, and impacts on table structure understanding tasks in llms}.
\newblock \emph{Preprint}, arXiv:2310.10358.

\bibitem[{Wallace et~al.(2019)Wallace, Wang, Li, Singh, and Gardner}]{wallace-etal-2019-nlp}
Eric Wallace, Yizhong Wang, Sujian Li, Sameer Singh, and Matt Gardner. 2019.
\newblock \href {https://doi.org/10.18653/v1/D19-1534} {Do {NLP} models know numbers? probing numeracy in embeddings}.
\newblock In \emph{Proceedings of the 2019 Conference on Empirical Methods in Natural Language Processing and the 9th International Joint Conference on Natural Language Processing (EMNLP-IJCNLP)}, pages 5307--5315, Hong Kong, China. Association for Computational Linguistics.

\bibitem[{Weber et~al.(2024)Weber, Fu, Anthony, Oren, Adams, Alexandrov, Lyu, Nguyen, Yao, Adams, Athiwaratkun, Chalamala, Chen, Ryabinin, Dao, Liang, Ré, Rish, and Zhang}]{weber2024redpajamaopendatasettraining}
Maurice Weber, Daniel Fu, Quentin Anthony, Yonatan Oren, Shane Adams, Anton Alexandrov, Xiaozhong Lyu, Huu Nguyen, Xiaozhe Yao, Virginia Adams, Ben Athiwaratkun, Rahul Chalamala, Kezhen Chen, Max Ryabinin, Tri Dao, Percy Liang, Christopher Ré, Irina Rish, and Ce~Zhang. 2024.
\newblock \href {https://arxiv.org/abs/2411.12372} {Redpajama: an open dataset for training large language models}.
\newblock \emph{Preprint}, arXiv:2411.12372.

\bibitem[{Winter(2013)}]{winter2013linearmodelslinearmixed}
Bodo Winter. 2013.
\newblock \href {https://arxiv.org/abs/1308.5499} {Linear models and linear mixed effects models in r with linguistic applications}.
\newblock \emph{Preprint}, arXiv:1308.5499.

\bibitem[{Zhang et~al.(2025)Zhang, Cao, Wei, Xu, and You}]{zhang2025tokenizationconstraintsllmsstudy}
Xiang Zhang, Juntai Cao, Jiaqi Wei, Yiwei Xu, and Chenyu You. 2025.
\newblock \href {https://arxiv.org/abs/2505.14178} {Tokenization constraints in llms: A study of symbolic and arithmetic reasoning limits}.
\newblock \emph{Preprint}, arXiv:2505.14178.

\end{thebibliography}


\appendix

\section{Experimental Setup} 

\subsection{Model Configuration} 
\label{sec:Model_Configuration}
We evaluate a diverse set of nine LLMs covering a range of sizes, licensing models (open-source and closed-source), model families (both small and large), and pretraining corpora (including multilingual models). This diversity allows us to examine how different architectures and training data influence performance on arithmetic tasks across scripts and typography. Models known or presumed to have at least 70B parameters are classified as \emph{large}, and the remainder as \emph{small}.

Since many of the output tokens for low-resource scripts are rare, we experimented with four temperature values (0.3, 0.7, 1.0, 1.5) to control the model’s sampling randomness and improve its ability to generate these infrequent tokens. Testing on a 30\% subset of the evaluation set for the large models showed that a temperature of 0.7 offered the best balance between faithful generation and sufficient exploration for correctly producing rare numeral symbols.

For the formatted versions of the tasks, a lower temperature of 0.3 was found to be more effective, as it encourages more deterministic outputs, helping the models reproduce the specified numeric formatting and arithmetic operators accurately.

\begin{table}[h!]
\centering
\begin{tabular}{l|c|c}
\hline
\textbf{Model} & \textbf{Size} & \textbf{Source} \\
\hline
\texttt{claude-sonnet-4-5-20250929} & - & Closed \\
\texttt{gpt-4o-2024-08-06} & - & Closed\\
\texttt{gemini-2.5-pro} & - & Closed \\
\texttt{Llama-3.3-70B-Instruct} & 70B & Open\\
\texttt{Llama-3.1-8B-Instruct} & 8B & Open\\
\texttt{Olmo-3-7B-Instruct} & 7B & Open\\
\texttt{bloomz-7b1} & 7B & Open\\
\texttt{gemma-3-4b-it} & 4B & Open\\
\texttt{Qwen3-4B-Instruct-2507} & 4B & Open\\
\hline
\end{tabular}
\caption{Summary of language models evaluated in this study.}
\label{tab:llm_summary}
\end{table}

\subsection{Evaluation}
For all tasks except Script Identification, model outputs are evaluated based on exact matching with the ground truth. In our prompts, we explicitly specify the desired output format, typically as \texttt{Answer: \$ANSWER}. An output is considered correct if the answer appears in this format anywhere within the model’s response. For the Translation task specifically, a response is counted as correct only if the translated numeral string exactly matches the target HA numeral string.
We also record cases where the model produces the correct numeric value but does not follow the specified format. These instances are treated separately and reported in the error analysis (\Cref{sec:error_analysis}), as they indicate that the model understands the underlying arithmetic or numeral translation but fails to adhere to the formatting instructions.

\section{Dataset}
\label{sec:dataset_generation}
We initially generated a set of 500 arithmetic expressions, with 125 questions per arithmetic operation. After a preliminary evaluation using basic LLMs, we removed 109 questions whose operands contained very large numbers causing all evaluated models to fail.

To translate expressions into various numeral scripts, we first consolidated digit mappings, decimal markers, and formatting conventions from multiple multilingual sources, including \href{https://www.omniglot.com/}{Omniglot}, \href{https://en.wikipedia.org/wiki/List_of_numeral_systems}{Wikipedia}, \href{https://r12a.github.io/}{r12a}, \href{https://www.unicode.org/cldr/charts/47/supplemental/languages_and_scripts.html}{Unicode} and \href{https://www.localeplanet.com/icu/decimal-symbols.html}{LocalePlanet}. In many scripts, we could not identify standardized mathematical operator symbols; therefore, we used word forms for arithmetic operators. To verify that this choice did not affect model behavior, we evaluated Claude Sonnet~4.5 on Hindu–Arabic numerals using both symbolic operators and their English word equivalents and observed identical performance. Based on this result, we consistently used word forms for all operators in our multiscript evaluations. Translations of both numerals and accompanying text were verified using a combination of native users of the script, Google Translate, and LLM-based judges (Claude Sonnet v4 and GPT-4). A translation was accepted if at least two of these four sources judged it to be correct. All native speakers consulted were friends or colleagues who participated voluntarily, provided informed consent, and received no monetary compensation for their contributions. Each annotator was shown numerical expressions written in a script they were familiar with, along with their equivalent representations in the HA script, and was asked to label whether each expression was correct. Annotators were also asked to assess the correctness of the prompts, in their respective scripts, used in this study.

For formatting variations, a Python script automatically generated six alternative versions of each arithmetic expression, applying the appropriate decimal markers and digit grouping conventions. All generated expressions were manually verified for correctness. Following this conversion, an additional 55 questions were removed because their operands could not be consistently classified into one of the six predefined formatting variants, leaving 336 expressions for the final analysis.

\section{Multiscript Numerical Reasoning for Smaller Language Models}
\label{sec:multilingual_small_llm}

\subsection{Distribution in Pretraining Corpora}
\label{sec:script_corpus_dist}

To quantify the representation of different numeral scripts in large-scale training data, we compute the average proportion of digit occurrences for each script used in this study across four widely used pretraining corpora: PILE \citep{gao2020pile800gbdatasetdiverse}, RedPajama \citep{weber2024redpajamaopendatasettraining}, OSCAR \citep{ortizsuarez}, and FineWeb2 \citep{penedo2025fineweb2pipelinescale}. For each corpus, we randomly sample 500k documents and extract all numeric digits appearing in the text. Each extracted number is then classified according to the numeral scripts considered in our experiments, with all remaining numerals grouped under an \emph{other} category. We then compute the proportion of digit occurrences attributed to each script and average these proportions across all samples. The reported percentages sum to 99.1\%, with the remaining 0.9\% corresponding to numerals from scripts not covered in this study. This analysis provides a coarse but informative estimate of the relative prevalence of different numeral scripts in contemporary pretraining data, helping contextualize the downstream performance trends observed in our experiments.

\subsection{Script Identification and Translation}
\label{sec:script_identification_small_llm}
In contrast to the large language models, the smaller models exhibit very low performance on both the script identification and translation tasks (\Cref{tab:small_llm_name_the_script_results} and \Cref{tab:small_llm_translation_results}, respectively). Most small models are unable to reliably recognize numeral scripts, often producing incorrect or partially related language names instead. Despite this overall low accuracy, the same strong correlation between performance on the two tasks observed in larger models is present, indicating that even small models share a dependency between script recognition and numeric translation, though at a much lower baseline level.

\begin{table}[]
\centering
\resizebox{\columnwidth}{!}{%
\begin{tabular}{lccccc}
\toprule

\textbf{Script} & \makecell{\textbf{Llama-3.1}\\ \textbf{8B}} & \makecell{\textbf{Bloomz}\\ \textbf{7B}} & \makecell{\textbf{Olmo-3}\\ \textbf{7B}} & \makecell{\textbf{Qwen-3}\\ \textbf{4B}}  & \makecell{\textbf{Gemma-3}\\ \textbf{4B}} \\
\midrule

Adlam & 0.0 & 0.0 & 0.0 & 0.0 & 0.0 \\
Balinese  & 0.1 & 0.0 & 0.0 & 0.0 & 0.1 \\
Bengali & \highlight{0.5} & 0.1 & 0.3 & 0.03 & \highlight{0.9} \\
Chinese\footnotesize{*} & 0.3 & 0.1 & 0.1 & 0.03 & 0.47 \\
Devanagari & \highlight{1.0} & 0.03 & \highlight{0.5} & 0.0 & 1.0 \\
Gujarati & \highlight{0.53} & 0.01 & 0.0 & 0.0 & 0.43 \\
Javanese & 0.2 & 0.0 & 0.0 & 0.0 & 0.1 \\
Kannada & \highlight{1.0} & 0.03 & 0.1 & 0.0 & 0.0 \\
Khmer & \highlight{1.0} & 0.0 & 0.0 & 0.1 & \highlight{1.0} \\
Lao  & \highlight{1.0} & 0.0 & 0.0 & 0.0 & 0.1\\
Malayalam & \highlight{1.0} & 0.0 & 0.17 & 0.0 & 0.27\\
Myanmar & \highlight{0.67} & 0.0 & 0.27 & 0.03 & 0.4\\
N'ko & \highlight{1.0} & 0.0 & 0.0 & 0.0 & 0.0\\
Ol Chiki  & 0.03 & 0.0 & 0.0 & 0.0 & 0.17 \\
Oriya & \highlight{1.0} & 0.0 & 0.3 & 0.0 & 0.03 \\
Osmanya & 0.0 & 0.0 & 0.0 & 0.0 & 0.1 \\
Perso-Arabic & 0.4 & 0.17 & 0.3 & 0.0 & \highlight{0.77}\\
Tamil & \highlight{1.0} & 0.0 & 0.0 & 0.0 & 0.47\\
Telugu & \highlight{0.83} & 0.03 & 0.17 & 0.0 & 0.27\\
Thai & \highlight{0.97} & 0.0 & 0.03 & 0.03 & \highlight{0.93} \\

\midrule
\textbf{Avg. Accuracy} & \textbf{0.62} & \textbf{0.02} & \textbf{0.11} & \textbf{0.01} & \textbf{0.38} \\
\bottomrule
\end{tabular}%
}
\caption{Accuracy on the script identification task for small language models. Cells highlighted in yellow indicate cases where model accuracy is 0.5 or higher.\footnotesize{ *Simplified Chinese script}}
\label{tab:small_llm_name_the_script_results}
\end{table}

\begin{table}[]
\centering
\resizebox{\columnwidth}{!}{%
\begin{tabular}{lccccc}
\toprule

\textbf{Script} & \makecell{\textbf{Llama-3.1}\\ \textbf{8B}} & \makecell{\textbf{Bloomz}\\ \textbf{7B}} & \makecell{\textbf{Olmo-3}\\ \textbf{7B}} & \makecell{\textbf{Qwen-3}\\ \textbf{4B}}  & \makecell{\textbf{Gemma-3}\\ \textbf{4B}} \\
\midrule

Adlam & 0.0 & 0.0 & 0.0 & 0.0 & 0.0 \\
Balinese  & 0.1 & 0.0 & 0.0 & 0.0 & 0.1 \\
Bengali & \highlight{0.5} & 0.0 & 0.1 & 0.03 & \highlight{0.9} \\
Chinese\footnotesize{*} & 0.3 & 0.0 & 0.0 & 0.03 & 0.47 \\
Devanagari & \highlight{1.0} & 0.1 & \highlight{0.5} & 0.03 & 1.0 \\
Gujarati & \highlight{0.53} & 0.01 & 0.0 & 0.17 & 0.43 \\
Javanese & 0.2 & 0.0 & 0.0 & 0.0 & 0.1 \\
Kannada & \highlight{1.0} & 0.03 & 0.1 & 0.0 & 0.0 \\
Khmer & \highlight{1.0} & 0.0 & 0.0 & 0.03 & \highlight{1.0} \\
Lao  & \highlight{1.0} & 0.0 & 0.0 & 0.0 & 0.0\\
Malayalam & \highlight{1.0} & 0.0 & 0.03 & 0.0 & 0.27\\
Myanmar & \highlight{0.67} & 0.0 & 0.0 & 0.03 & 0.4\\
N'ko & \highlight{1.0} & 0.0 & 0.0 & 0.0 & 0.0\\
Ol Chiki  & 0.03 & 0.0 & 0.0 & 0.0 & 0.0 \\
Oriya & \highlight{1.0} & 0.0 & 0.0 & 0.0 & 0.03 \\
Osmanya & 0.0 & 0.0 & 0.0 & 0.0 & 0.1 \\
Perso-Arabic & 0.4 & 0.0 & 0.0 & 0.0 & \highlight{0.77}\\
Tamil & \highlight{1.0} & 0.0 & 0.0 & 0.0 & 0.0\\
Telugu & \highlight{0.83} & 0.0 & 0.0 & 0.0 & 0.0\\
Thai & \highlight{0.97} & 0.0 & 0.0 & 0.03 & \highlight{0.93} \\

\midrule
\textbf{Avg. Accuracy} & \textbf{0.62} & \textbf{0.01} & \textbf{0.04} & \textbf{0.01} & \textbf{0.33} \\
\bottomrule
\end{tabular}%
}
\caption{Accuracy on the translation task for smaller language models. Cells highlighted in yellow indicate cases where model accuracy is 0.5 or higher.\footnotesize{ *Simplified Chinese script}}
\label{tab:small_llm_translation_results}
\end{table}

\subsection{Arithmetic Computation}
\label{sec:script_arithmetic_computation}

\Cref{tab:script_language_mapping} shows the script-to-language mapping used for prompting in this study. Note that multiple languages can share the same script; for each script, we selected the language we considered the most commonly used.

\begin{table}[]
\centering
\begin{tabular}{ll}
\toprule

\textbf{Script} & \textbf{Language} \\
\midrule
 
Hindu-Arabic & English \\
Adlam & Fulani \\
Balinese & Balinese \\
Bengali & Bengali \\
Chinese (Simplified) & Mandarin \\
Devanagari & Hindi \\
Gujarati & Gujarati \\
Javanese & Javanese \\
Kannada & Kannada \\
Khmer & Khmer \\
Lao  & Lao \\
Malayalam & Malayalam \\
Myanmar & Burmese \\
N'ko & Maninka \\
Ol Chiki  & Santali \\
Oriya & Odia \\
Osmanya & Somali \\
Perso-Arabic & Persian \\
Tamil & Tamil \\
Telugu & Telugu \\
Thai & Thai \\

\bottomrule
\end{tabular}%

\caption{Numeral scripts considered in this work and the corresponding languages used for prompting.}
\label{tab:script_language_mapping}
\end{table}

\begin{figure*}[]
    \centering
    \includegraphics[width=0.8\linewidth]{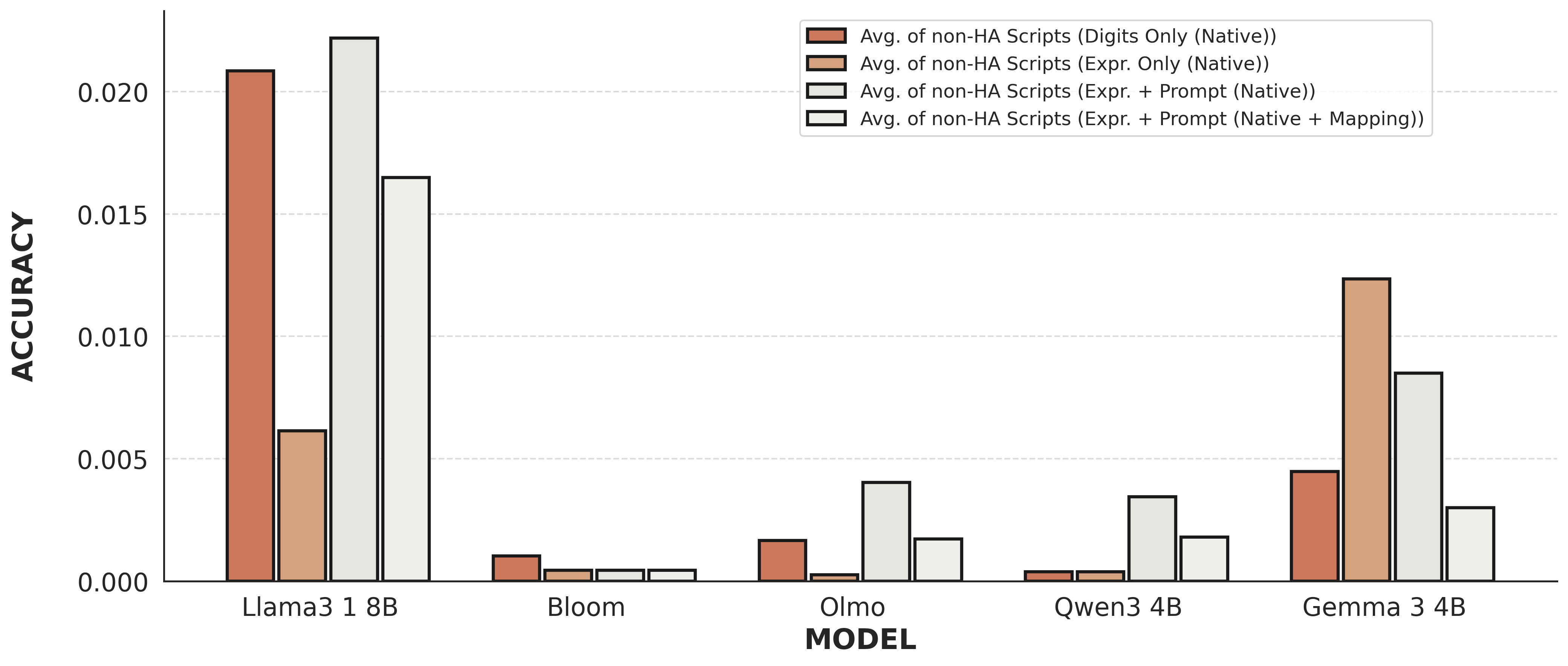}
    \caption{Comparison of model accuracy on the arithmetic task. For each language model, the bars shows the mean accuracy across all non-Hindu-Arabic (non-HA) scripts for varying prompts.}
    \label{fig:small_llm_org_v_all_script_accuracy}
\end{figure*}

The performance of all smaller models is very low, regardless of the prompting strategy as shown in \Cref{fig:small_llm_org_v_all_script_accuracy}. In many cases, these models exhaust their maximum generation limits before producing a valid prediction. When outputs are generated, they often exhibit confusion at the language-name level—e.g., producing phonetically or orthographically similar languages—rather than correctly identifying the numeral script. Given this consistently poor performance, we restrict the remaining analysis to the larger models.

\subsection{Prompt Examples}
\label{sec:multilingual_prompt_examples}

\begin{otherlanguage}{chinese-simplified}
\begin{CodeBox}[Script Identification]

    \Role{system}{What script is used to represent the numerals in the equation}
    
    \Role{user}{{\hindifont{९१३७५३०}} - {\hindifont{३६५१८}}? Only return the name of the script without any additional text}
    
    \ExpectedOutput{Devanagari}
\end{CodeBox}
\end{otherlanguage}

\begin{otherlanguage}{chinese-simplified}
\begin{CodeBox}[Translation]

    \Role{system}{Translate the given number to Hindu-Arabic numerals and respond ONLY with the translated number. The output should be of the form: \texttt{``Answer: \$ANSWER''} (without quotes) where \$ANSWER is the translated number.}
    
    \Role{user}{{\kannadafont{೮೯೪೨೮}}}
    
    \ExpectedOutput{Answer: 89428}
\end{CodeBox}
\end{otherlanguage}

\begin{otherlanguage}{chinese-simplified}
\begin{CodeBox}[Digits Only (Native)]
    \Role{system}{Compute and respond ONLY with the answer. The output should be of the form: ``Answer: \$ANSWER'' (without quotes) where \$ANSWER is the answer to the problem. Ensure the answer is in the same script as the numbers in the question.}
    
    \Role{user}{Round the answer to an integer.\\ 三百八十二万六千九百九十五 \hspace{0.5mm} divided by \hspace{0.5mm} 五十四万九千二百零七}
    
    \ExpectedOutput{Answer: 七}
\end{CodeBox}
\end{otherlanguage}

\begin{otherlanguage}{chinese-simplified}
\begin{CodeBox}[Expression Only (Native)]

    \Role{system}{Compute and respond ONLY with the answer. The output should be of the form: ``Answer: \$ANSWER'' (without quotes) where \$ANSWER is the answer to the problem. Ensure the answer is in the same script as the numbers in the question.}
    
    \Role{user}{Round the answer to an integer.\\ 三百八十二万六千九百九十五除以五十四万九千二百零七}
    
    \ExpectedOutput{Answer: 七}
    
\end{CodeBox}
\end{otherlanguage}

\begin{otherlanguage}{chinese-simplified}
\begin{CodeBox}[Expression + Prompt (Native)]

    \Role{system}{Compute and respond ONLY with the answer. The output should be of the form: ``Answer: \$ANSWER'' (without quotes) where \$ANSWER is the answer to the problem. Ensure the answer is in the same script as the numbers in the question.}
    
    \Role{user}{将答案四舍五入到整数。\\三百八十二万六千九百九十五除以五十四万九千二百零七}
    
    \ExpectedOutput{Answer:七}
    
\end{CodeBox}
\end{otherlanguage}

\begin{otherlanguage}{chinese-simplified}
\begin{CodeBox}[Expression + Prompt (Native + Mapping)]

    \Role{system}{Compute and respond ONLY with the answer. The output should be of the form: ``Answer: \$ANSWER'' (without quotes) where \$ANSWER is the answer to the problem. Ensure the answer is in the same script as the numbers in the question, with the mapping between the script’s numerals and Latin numerals provided as reference.}
    
    \Role{user}{
        [〇:0, 一:1, 二:2, 三:3, 四:4, 五:5, 六:6, 七:7, 八:8, 九:9,十:10, 百:100, 千:1000, 万:10000, 亿:100000]\\ 将答案四舍五入到整数。\\ 三百八十二万六千九百九十五除以五十四万九千二百零七}
        
    \ExpectedOutput{Answer:七}
    
\end{CodeBox}
\end{otherlanguage}

\section{Arithmetic Reasoning with Formatted Numbers}

\subsection{Distribution in Pretraining Corpora}
\label{sec:format_corpus_dist}
To measure the representation of the numeric formatting conventions studied in this work within large-scale pretraining data, we analyze four widely used corpora: PILE, RedPajama, OSCAR, and FineWeb2. For each corpus, we randomly sample 500k documents and extract numeric strings using regular expressions tailored to match each formatting pattern. To avoid ambiguity, we discard any numeric string that can be mapped to more than one format (e.g., \texttt{125.3}, which could correspond to F1, F4, or F6). We then compute the proportion of occurrences for each format and average these proportions across all samples. This procedure yields an estimate of how frequently different numeric formatting conventions appear in large-scale pretraining data.

\subsection{Performance of Smaller Language Models}
\label{sec:formatted_small_llm}

\Cref{tab:small_llm_format_tokenization} shows the tokenization of input equations across the six formatting variants for the small LLMs evaluated in this study. LLama~3.1 8B uses the same tokenizer as LLama~3.3 70B and Gemma~3 uses the same tokenizer as Gemini~2.5 Pro depicted in \Cref{tab:format_tokenization}.

\begin{figure}[H]
    \centering
    \includegraphics[width=1\linewidth]{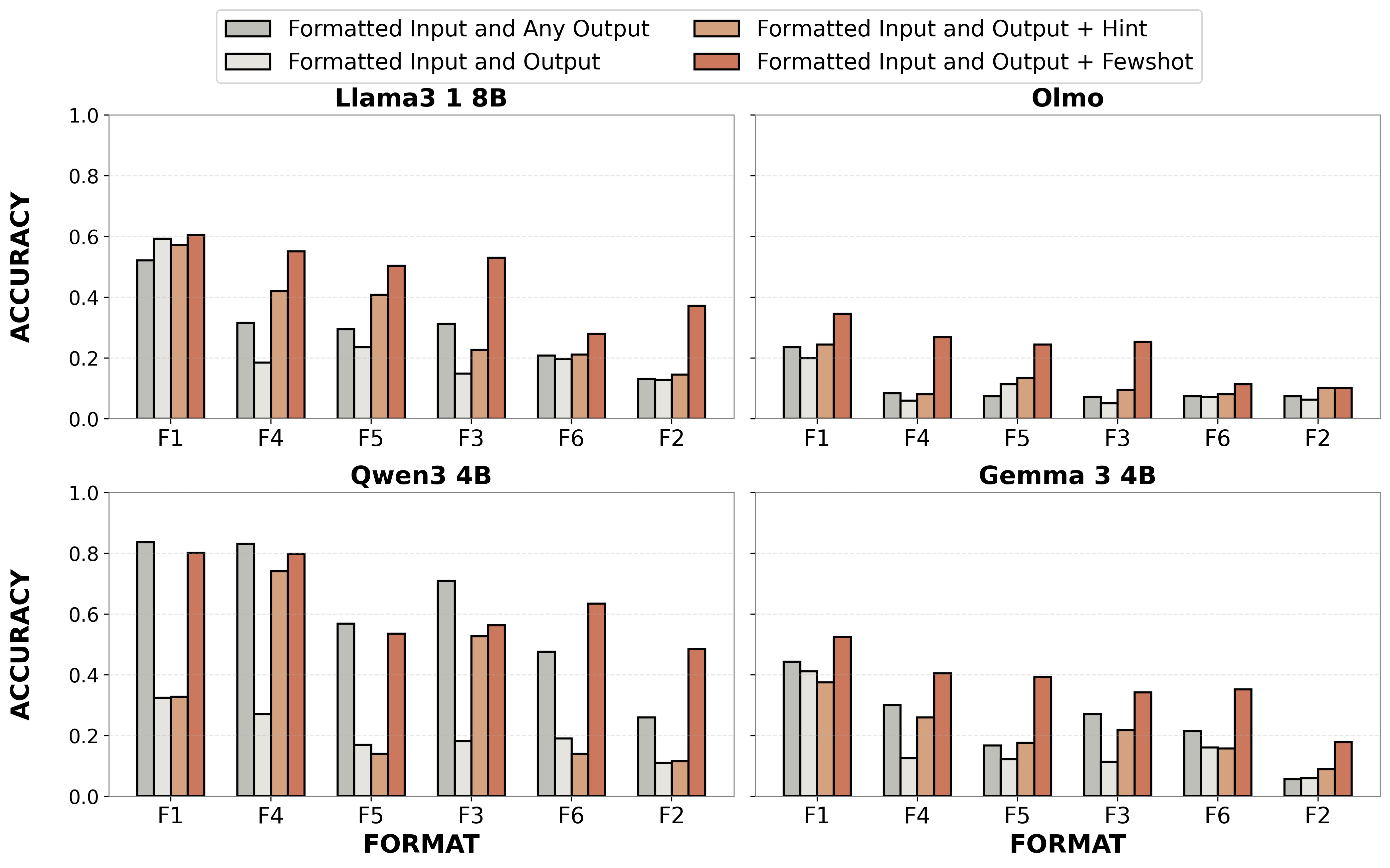}
    \caption{Average accuracy per format variant for each small language model across different prompting strategies. Each subplot represents a single model, and bars indicate the performance of four prompting strategies. Formats are sorted by their overall average accuracy.}
    \label{fig:small_llm_format_all_prompts}
\end{figure}

As illustrated in \Cref{fig:small_llm_format_all_prompts}, format F1 consistently yields the highest accuracy across all tested models, while F2 presents the greatest challenge, often resulting in the lowest success rates regardless of the prompting strategy. Among the four models evaluated, Qwen3 4B demonstrates the highest peak performance, comparable to that of large language models. Conversely, Olmo shows more limited numerical reasoning capabilities, failing to surpass 0.4 accuracy even under the most favorable prompting conditions.

\begingroup
\setlength{\tabcolsep}{2pt} 
\renewcommand{\arraystretch}{0.9} 

\begin{table*}[]
\centering
\small
\setlength{\extrarowheight}{10pt}
\resizebox{\textwidth}{!}{%
\begin{tabular}{|c|l|c|c|c|}
\toprule

\textbf{Format} & \textbf{Sample} & \textbf{Bloomz} & \textbf{Olmo-3} & \textbf{Qwen~3}\\
\midrule

F1 & \ttfamily{922,436.38 + 4,359} 
& \token{922}\token{,}\token{436}\token{.}\token{38}\token{ +}\token{ 4}\token{,}\token{359} 
& \token{922}\token{,}\token{436}\token{.}\token{38}\token{ +}\token{ }\token{4}\token{,}\token{359} 
& \token{9}\token{2}\token{2}\token{,}\token{4}\token{3}\token{6}\token{.}\token{3}\token{8}\token{ +}\token{ }\token{4}\token{,}\token{3}\token{5}\token{9} \\

F2 & \ttfamily{922.436,38 + 4.359}
& \token{922}\token{.}\token{436}\token{,}\token{38}\token{ +}\token{ 4}\token{.}\token{359} 
& \token{922}\token{.}\token{436}\token{,}\token{38}\token{ +}\token{ }\token{4}\token{.}\token{359} 
& \token{9}\token{2}\token{2}\token{.}\token{4}\token{3}\token{6}\token{,}\token{3}\token{8}\token{ +}\token{ }\token{4}\token{.}\token{3}\token{5}\token{9} \\

F3 & \ttfamily{922\textvisiblespace436,38 + 4\textvisiblespace359}
& \token{922}\token{\textvisiblespace}\token{436}\token{,}\token{38}\token{ +}\token{ 4}\token{\textvisiblespace}\token{359} 
& \token{922}\unknowntokene{$^1$}\unknowntokene{$^2$}\token{436}\token{,}\token{38}\token{ +}\token{ }\token{4}\unknowntokene{$^1$}\unknowntokene{$^2$}\token{359} 
& \token{9}\token{2}\token{2}\unknowntokene{$^1$}\unknowntokene{$^2$}\token{4}\token{3}\token{6}\token{,}\token{3}\token{8}\token{ +}\token{ }\token{4}\unknowntokene{$^1$}\unknowntokene{$^2$}\token{3}\token{5}\token{9} \\

F4 & \ttfamily{922\textvisiblespace436.38 + 4\textvisiblespace359}
& \token{922}\token{\textvisiblespace}\token{436}\token{.}\token{38}\token{ +}\token{ 4}\token{\textvisiblespace}\token{359} 
& \token{922}\unknowntokene{$^1$}\unknowntokene{$^2$}\token{436}\token{.}\token{38}\token{ +}\token{ }\token{4}\unknowntokene{$^1$}\unknowntokene{$^2$}\token{359} 
& \token{9}\token{2}\token{2}\unknowntokene{$^1$}\unknowntokene{$^2$}\token{4}\token{3}\token{6}\token{.}\token{3}\token{8}\token{ +}\token{ }\token{4}\unknowntokene{$^1$}\unknowntokene{$^2$}\token{3}\token{5}\token{9} \\

F5 & \ttfamily{922'436,38 + 4'359}
& \token{922}\token{'}\token{436}\token{,}\token{38}\token{ +}\token{ 4'}\token{359} 
& \token{922}\token{'}\token{436}\token{,}\token{38}\token{ +}\token{ }\token{4}\token{'}\token{359} 
& \token{9}\token{2}\token{2}\token{'}\token{4}\token{3}\token{6}\token{,}\token{3}\token{8}\token{ +}\token{ }\token{4}\token{'}\token{3}\token{5}\token{9} \\

F6 & \ttfamily{9,22,436.38 + 4,359}
& \token{9}\token{,}\token{22}\token{,}\token{436}\token{.}\token{38}\token{ +}\token{ 4}\token{,}\token{359} 

& \token{9}\token{,}\token{22}\token{,}\token{436}\token{.}\token{38}\token{ +}\token{ }\token{4}\token{,}\token{359} 
& \token{9}\token{,}\token{2}\token{2}\token{,}\token{4}\token{3}\token{6}\token{.}\token{3}\token{8}\token{ +}\token{ }\token{4}\token{,}\token{3}\token{5}\token{9} \\

\bottomrule
\end{tabular}
}
\caption{Tokenization of input equations across the six formatting variants for various small LLMs.}
\label{tab:small_llm_format_tokenization}
\end{table*}

\endgroup

The figure also highlights a ``formatting penalty'' for smaller models, where requiring a specific output format drastically reduces accuracy compared to allowing any output format, particularly for Qwen3~4B and Llama3.1~8B. Introducing few-shot examples effectively mitigates this bottleneck, consistently serving as the most robust prompting strategy across all models and formats. In many cases, such as with Gemma~3~4B and Qwen3~4B, few-shot prompting not only recovers accuracy lost to formatting constraints but even surpasses the "Any Output" baseline, emphasizing the importance of in-context examples for guiding smaller architectures through complex numerical transformations.

\subsection{Prompt Examples}
\label{sec:format_prompt_examples}

\begin{otherlanguage}{chinese-simplified}
\begin{CodeBox}[Formatted Input and Any Output]

    \Role{system}{Compute and respond ONLY with the answer. The output should be of the form: ``Answer: \$ANSWER'' (without quotes) where \$ANSWER is the answer to the problem.}
    
    \Role{user}{Round the answer to three decimal places.\\ 
    22\textvisiblespace436,447 plus 4\textvisiblespace359}
    
    \ExpectedOutput{Answer: 26795.447}
\end{CodeBox}
\end{otherlanguage}

\begin{otherlanguage}{chinese-simplified}
\begin{CodeBox}[Formatted Input and Output]

    \Role{system}{Compute and respond ONLY with the answer. The output should be of the form: ``Answer: \$ANSWER'' (without quotes) where \$ANSWER is the answer to the problem. Ensure the answer has the same formatting as the numbers in the question.}
    
    \Role{user}{Round the answer to three decimal places.\\ 
    22\textvisiblespace436,447 plus 4\textvisiblespace359}
    
    \ExpectedOutput{Answer: 26\textvisiblespace795,447}
    
\end{CodeBox}
\end{otherlanguage}

\begin{otherlanguage}{chinese-simplified}
\begin{CodeBox}[Formatted Input and Output + Hint]

    \Role{system}{Compute and respond ONLY with the answer. The output should be of the form: ``Answer: \$ANSWER'' (without quotes) where \$ANSWER is the answer to the problem. Ensure the answer has the same formatting as the numbers in the question with decimal markers and grouping separators as mentioned.}
    
    \Role{user}{The decimal marker used is `,' and the grouping separator is `\textvisiblespace'.\\[1em]
    Round the answer to three decimal places.\\ 
    22\textvisiblespace436,447 plus 4\textvisiblespace359}
    
    \ExpectedOutput{Answer: 26\textvisiblespace795,447}
    
\end{CodeBox}
\end{otherlanguage}

\begin{otherlanguage}{chinese-simplified}
\begin{CodeBox}[Formatted Input and Output + Fewshot]

    \Role{system}{Compute and respond ONLY with the answer. The output should be of the form: ``Answer: \$ANSWER'' (without quotes) where \$ANSWER is the answer to the problem. Ensure the answer has the same formatting as the numbers in the question.}
    
    \Role{user}{Here are some examples:\\ 
    Example 1:\\ 
    Round the answer to three decimal places.\\
    4\textvisiblespace958,155 multiplied by 93,2 \\
    Answer: 462\textvisiblespace100,046\\[1em]
    Example 2:\\
    Round the answer to three decimal places.\\
    9\textvisiblespace628\textvisiblespace240 divided by 4\textvisiblespace847\\ 
    Answer: 1\textvisiblespace986,433 \\[1em]
    Round the answer to three decimal places.\\ 
    22\textvisiblespace436,447 plus 4\textvisiblespace359}
    
    \ExpectedOutput{Answer: 26\textvisiblespace795,447}
    
\end{CodeBox}
\end{otherlanguage}

\subsection{Error analysis: Impact of Numeral Formats on Model Reasoning}
\label{sec:error_analysis}

To better understand the failure modes of large language models (LLMs) when processing diverse numerical representations, we perform an error analysis across three primary categories: 

\begin{itemize}
    \item \textit{Instruction Errors}: In most prompts, we require the model to follow specific output instructions. While minor deviations are permitted, such as omitting an explicit \texttt{ANSWER:} prefix, we expect models to adhere to constraints such as returning results in integer form or rounded to three decimal places. This category includes all errors arising from failure to follow these instructions.
    
    \item \textit{Arithmetic Errors}: This category captures cases where the model produces an incorrect numerical result, i.e., the arithmetic computation itself is incorrect, regardless of whether the output format is otherwise valid.
    
    \item \textit{Formatting Errors}: For most prompting strategies, we expect the model to reproduce the same numeric formatting as used in the input expression. Formatting errors occur when the model computes the correct numerical value but fails to express it using the required typographic conventions.

    \item \textit{No Output}: The model fails to produce a valid answer with or without reaching the maximum generation limit. In most of these cases, the model typically continues generating intermediate reasoning, repetitions, or unrelated text and exhausts the allowed token budget without emitting a final answer in the required format. 

\end{itemize}

\begin{figure*}[htbp]
    \centering
    \includegraphics[width=1\textwidth]{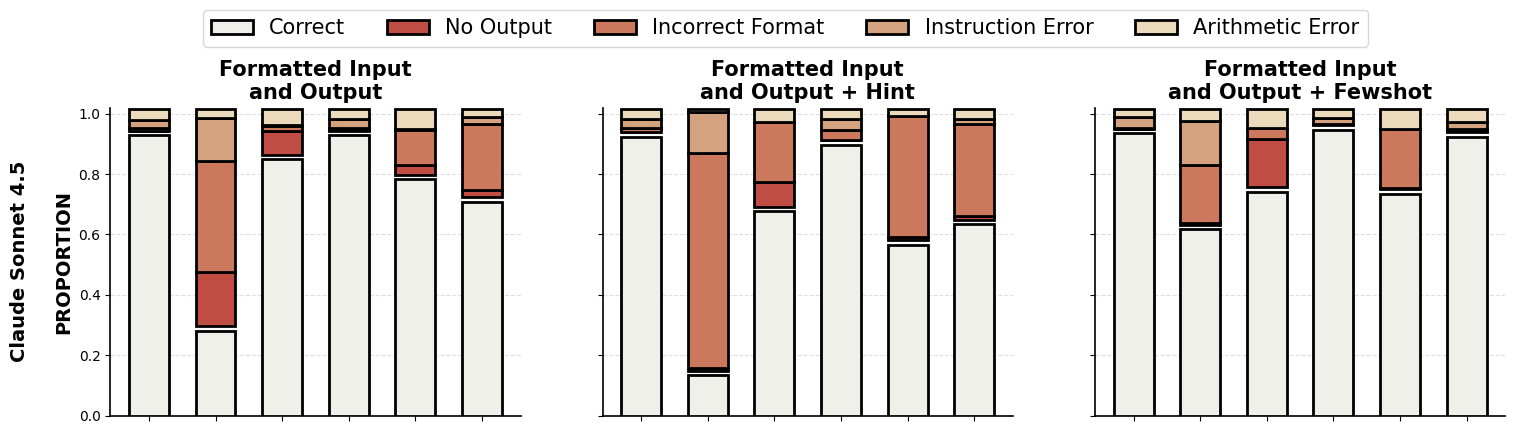}
    \vspace{0.1cm}
    
    \includegraphics[width=1\textwidth]{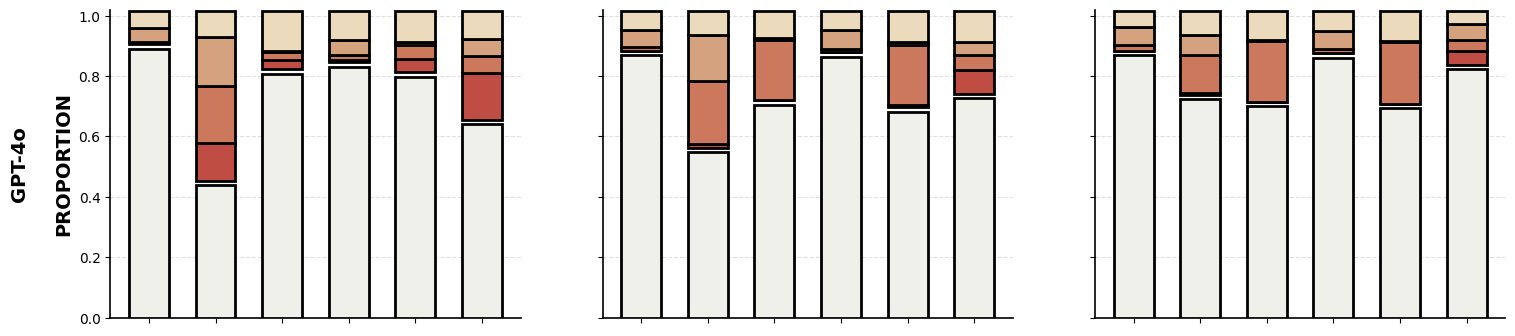}
    \vspace{0.1cm}

    \includegraphics[width=1\textwidth]{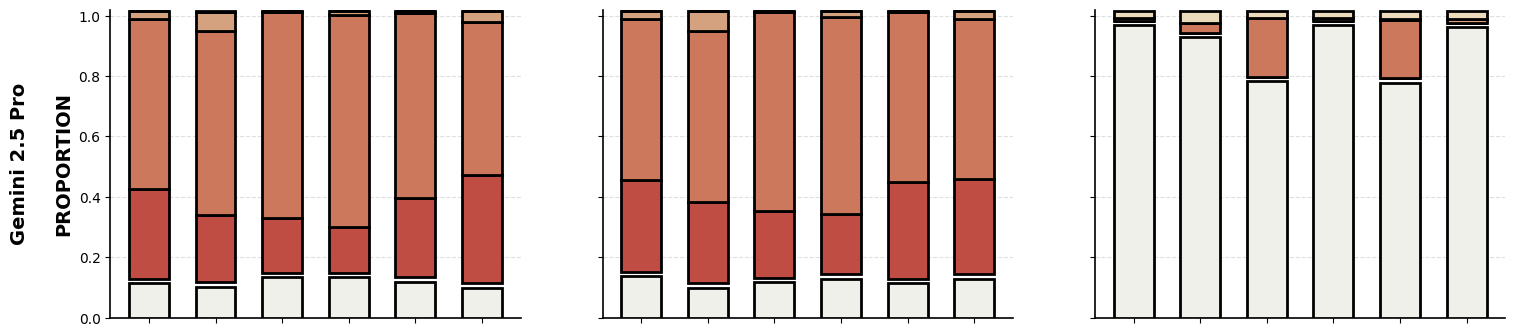}
    \vspace{0.1cm}
    
    \includegraphics[width=1\textwidth]{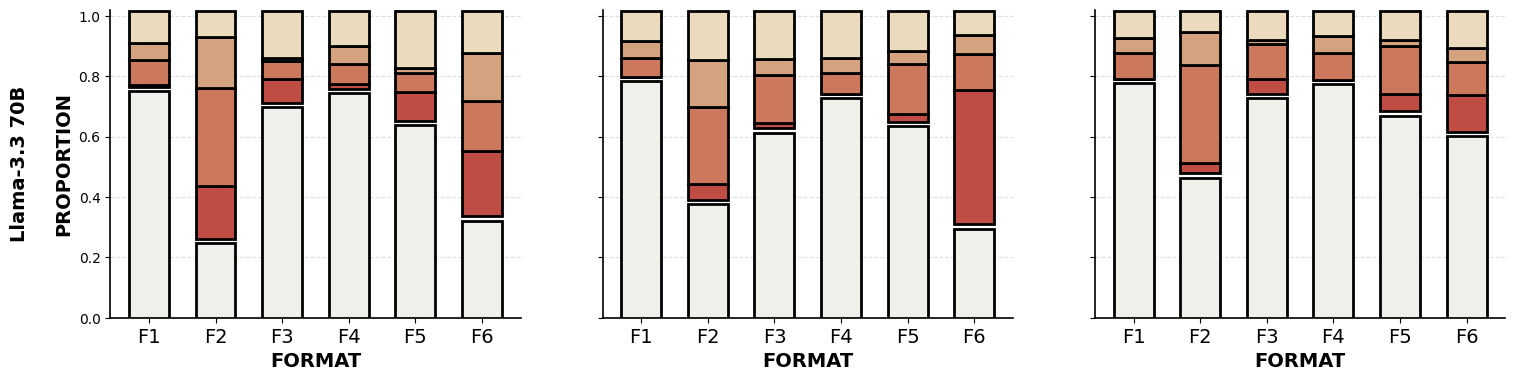}

    \caption{Distribution of error types across numeric formats (F1–F6) for different models and prompting strategies. Each stacked bar represents the proportion of correct answers, missing answers, and various error types for a given format.}
    \label{fig:format_error_analysis}
\end{figure*}

The distribution of these errors across numeral formats F1 through F6 for various LLMs is shown in \Cref{fig:format_error_analysis}.

\paragraph*{Impact of Prompting Strategies on Error Types}
Our analysis reveals that error distributions are highly sensitive to the prompting strategy employed.

\medskip
\noindent{Zero-Shot Sensitivity:} In the Formatted Input, Formatted Output setting, models frequently struggle with Instruction Errors and Incorrect Formats, particularly in formats like F2 and F6. This suggests that without explicit guidance, models often fail to parse complex or non-standard numerical representations.

\medskip
\noindent{The Power of Few-Shot Learning:} The introduction of few-shot samples (right-most subplot) significantly mitigates Instruction Errors across all models. For instance, in Gemini 2.5 Pro, the transition from zero-shot to few-shot almost entirely eliminates the Incorrect Format errors (red/orange bars), shifting the bottleneck toward core Arithmetic Errors.

\paragraph*{Model-Specific Error Trends}
Different model families exhibit unique failure modes when processing numerical data:

\medskip
\noindent{Claude Sonnet 4.5 and GPT-4o:} These models show a high degree of robustness. Their primary failures in more challenging formats (F2) are Arithmetic Errors, indicating that while they understand the required output format and the numerical system, they occasionally fail at the underlying calculation logic.

\medskip
\noindent{Llama-3.3 70B:} Llama exhibits a more distributed error profile. Even with few-shot samples, it maintains a significant proportion of Arithmetic Errors and Instruction Errors in formats F2 and F6, suggesting that its internal representation of these specific formats may be less robust than other models.

\paragraph*{Format-Dependent Difficulty}
F2, F3 and F6 formats consistently emerge as challenging formats across all models and strategies. These formats trigger the highest proportions of Instruction Errors, suggesting that the structural complexity of these formats makes it difficult for models to map the input to the desired output schema. In contrast, F1 and F4 seem to appear more natural to the models, with errors being almost exclusively Arithmetic. This implies that for these formats, the bottleneck is not the representation but the mathematical reasoning itself.

\end{document}